\documentclass[sigconf]{acmart}
\usepackage{times}
\usepackage{soul}
\usepackage{url}
\usepackage[utf8]{inputenc}
\usepackage{graphicx}
\usepackage{subfigure}
\usepackage{multirow}
\usepackage{amssymb}
\usepackage{amsmath}
\usepackage{amsfonts}
\usepackage{booktabs}
\usepackage{algorithm}
\usepackage{algorithmic}
\usepackage[font=small,labelfont=bf]{caption}

\renewcommand{\algorithmicrequire}{ \textbf{Input:}}
\renewcommand{\algorithmicoutput}{ \textbf{Output:}}
\renewcommand{\algorithmicensure}{ \textbf{Initialize:}}

\newtheorem{theorem}{Theorem}
\newtheorem{lemma}{Lemma}

\newtheorem{assumption}{Assumption}

\AtBeginDocument{
  \providecommand\BibTeX{{
    \normalfont B\kern-0.5em{\scshape i\kern-0.25em b}\kern-0.8em\TeX}}}

\setcopyright{acmcopyright}
\copyrightyear{2019}
\acmYear{2019}
\acmConference[November]{}{2019}{Beijing}
\begin{document}

\title{signADAM: Learning Confidences for Deep Neural Networks}
\acmSubmissionID{1273}

\author{Dong Wang}
\email{2947365183hk@gmail.com}
\authornote{Equal contribution.}
\affiliation{
  \institution{School of Artificial Intelligence, Xidian University}}

\author{Yicheng Liu}
\authornotemark[1]
\email{moooooore66@gmail.com}
\affiliation{
  \institution{School of Electronic Engineering, Xidian University}}

\author{Wenwo Tang}
\email{tww1507019@126.com}
\affiliation{
  \institution{School of Artificial Intelligence, Xidian University}}

\author{Fanhua Shang}
\email{fhshang@xidian.edu.cn}
\authornote{Corresponding author.}
\affiliation{
  \institution{School of Artificial Intelligence, Xidian University}}

\author{Hongying Liu}
\email{hyliu@xidian.edu.cn}
\affiliation{
  \institution{School of Artificial Intelligence, Xidian University}}

\author{Qigong Sun}
\email{xd_qigongsun@163.com}
\affiliation{
  \institution{School of Artificial Intelligence, Xidian University}}

\author{Licheng Jiao}
\email{lchjiao@mail.xidian.edu.cn}
\affiliation{
  \institution{School of Artificial Intelligence, Xidian University}}

\renewcommand{\shortauthors}{None}

\begin{abstract}
 In this paper, we propose a new first-order gradient-based algorithm to train deep neural networks. We first introduce the sign operation of stochastic gradients (as in sign-based methods, e.g., SIGN-SGD) into ADAM, which is called as signADAM. Moreover, in order to make the rate of fitting each feature closer, we define a confidence function to distinguish different components of gradients and apply it to our algorithm. It can generate more sparse gradients than existing algorithms do. We call this new algorithm signADAM++. In particular, both our algorithms are easy to implement and can speed up training of various deep neural networks. The motivation of signADAM++ is preferably learning features from the most different samples by updating large and useful gradients regardless of useless information in stochastic gradients. We also establish theoretical convergence guarantees for our algorithms. Empirical results on various datasets and models show that our algorithms yield much better performance than many state-of-the-art algorithms including SIGN-SGD, SIGNUM and ADAM. We also analyze the performance from multiple perspectives including the loss landscape and develop an adaptive method to further improve generalization. The source code is available at \url{ https://github.com/DongWanginxdu/signADAM-Learn-by-Confidence}.
\end{abstract}


\maketitle

\section{Introduction}
Deep neural networks (DNNs) are widely used in various fields of machine learning \cite{lecun2015deep} (e.g., speech recognition, visual object recognition and object detection) and achieve huge success in all these applications. Moreover, training DNNs is usually considered as a non-convex optimization problem. Stochastic gradient descent (SGD) is one of the most effective algorithms to train various DNNs. It is a method to minimize an objective function, which is parameterized by the model's parameters $\theta \!\in\! \mathbb{R}^{d}$ ($d$ is the number of the parameters in a model). SGD updates the parameters through the negative directions of the gradients of an objective function.

\begin{figure}[t]
\centering
\includegraphics[width=0.98\columnwidth]{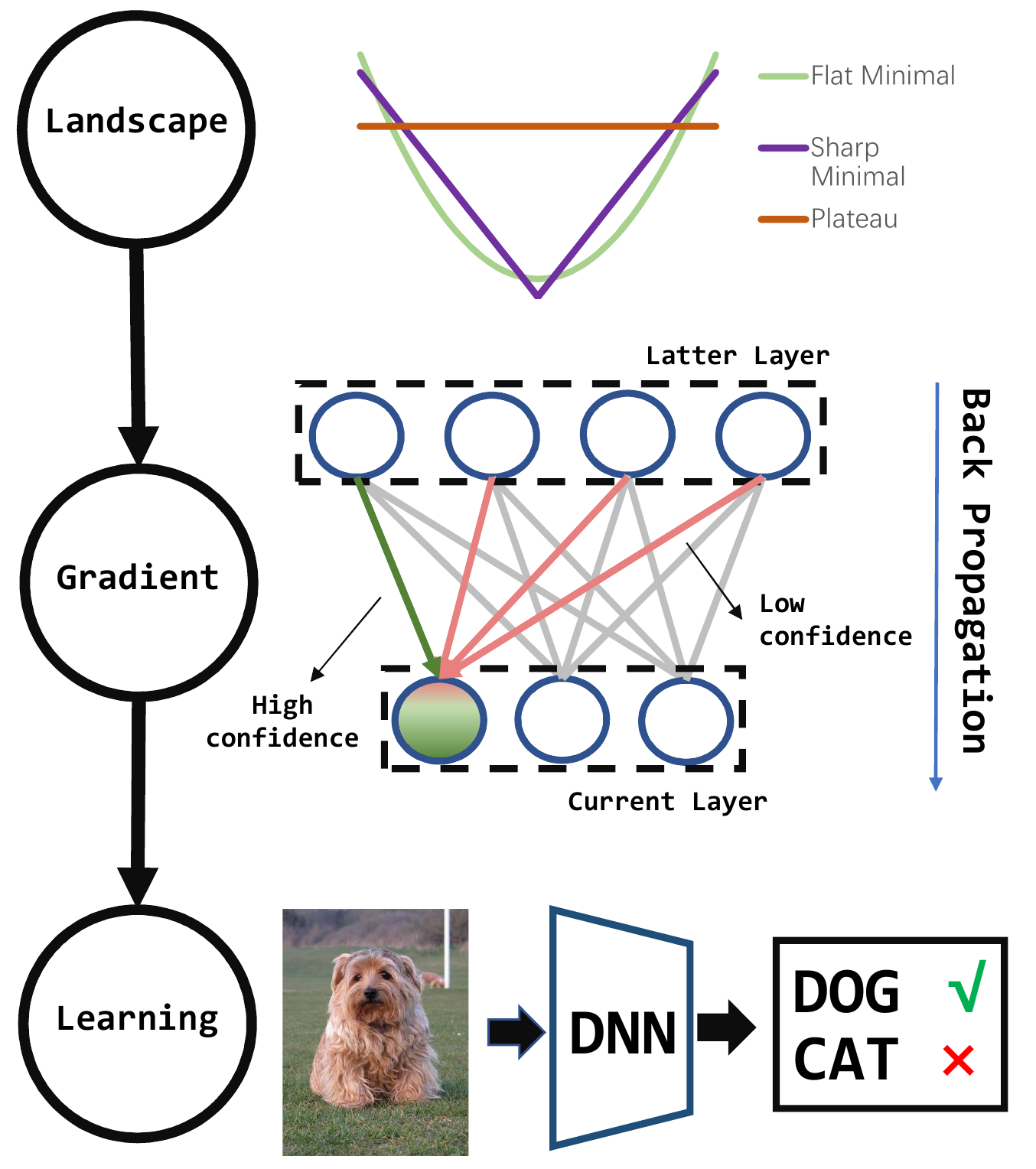}
\vspace{-3mm}

\caption{The connection between the loss landscape and learning. Loss landscape reflects the degree of feature learning. The gradients in different loss landscapes are various, and thus we give them different confidences by our defined confidence function in the training stage to enhance feature learning. A small gradient means that the corresponding feature is easy to learn so it should get a small confidence. An unbalanced feature learning leads to a sharp minimal. If the models learn features too unequally which is caused by updating too much or stopping learning, the algorithm would convergence to a very sharp minimal around a plateau.}
\label{flow_chart}
\end{figure}

SGD was first introduced by \cite{robbins1951stochastic}. In recent years, a lot of work so far has focused on the adaptive methods. For example, the first algorithm, which achieves satisfactory performance in these fields, is ADAGRAD \cite{duchi2011adaptive}, and ADAGRAD performs well in sparse settings. However, in the high dimensional setting, the vanishing gradient problem becomes serious in the training of DNNs. To alleviate this issue, several algorithms based on the adaptive methods have been developed, such as ADADELTA \cite{zeiler2012adadelta}, RMSPROP \cite{Tieleman2012} and ADAM \cite{kingma2014adam}. ADAM is one of the most popular optimization algorithms in deep learning. Based on the exponentially decaying average of past gradients, ADAM derives from the moment estimation and uses the first and second moments to compute individual learning rates for different parameters. However, the adaptive methods not only require the first moment estimate but also need the second moment estimate, which means extra computation amount. Recently, SIGN-SGD and SIGNUM \cite{bernstein2018signsgd} have been proposed. They applied the sign operation to SGD and its momentum variant, and also achieved a competitive performance in several benchmarks.

\citet{hinton2006fast} indicated that in some computer vision tasks especially image classification, the gradients mainly come from the incorrect samples and we can improve convergence rate by repeating sampling them. In other words, if we make the fitting rate of correct samples much slower than that of the incorrect samples, the network training will converge faster in fact. In practice, although adaptive methods have a faster convergence rate than SGD with momentum, SGD usually has a better performance on the test set than the others. We hold the view that the adaptive method gives the easy features a excessively fast fitting rate so that the models do not learn each useful feature of the dataset equally and reasonably. Because deep neural networks can not only learn the features that humans can easily understand, but also learn the features that are not easy to explain. So far, we have not been able to illustrate which features are really useful and which are useless. Inspired by the maximum entropy model \cite{attwell2001energy}, we tend to treat each feature equally. In terms of the difference of the fitting rate for each feature, the parameters which are used to extract and choose features should be not updated in a similar rate.

Motivated by all the above insights, the sign-based methods proposed by \cite{bernstein2018signsgd}, and the work \cite{balles2017dissecting}, we propose an efficient algorithm (called signADAM) instead of spending time in calculating the accurate moment information. We first integrate the sign operation to ADAM and hope to combine the advantages of both sign-based methods and ADAM. Furthermore, we introduce the confidence function, which is used to distinguish different components of gradients, as shown in Figure \ref{flow_chart}. Therefore, we need a larger confidence for large components of gradients than the others. We multiply the gradients by its confidence element-wise. Considering that the second moment estimate in signADAM is close to a constant, we remove it and apply the confidence function to its gradients from mini-batch samples. Then we get a new algorithm called signADAM++.

Above all, we first put forward a sign-based method by applying the sign operation into ADAM. And by introducing the confidence function, signADAM is developed into signADAM++. It completely suits our new framework. In particular, our framework explained that the confidence function can help to distinguish which features are necessary and urgent to learn. Besides, this also meets biological principles. We also analyze the results by the loss landscape and reveal the relationship between feature learning and loss landscape.

\begin{figure*}[t]
\centering
  \includegraphics[width=0.989\textwidth]{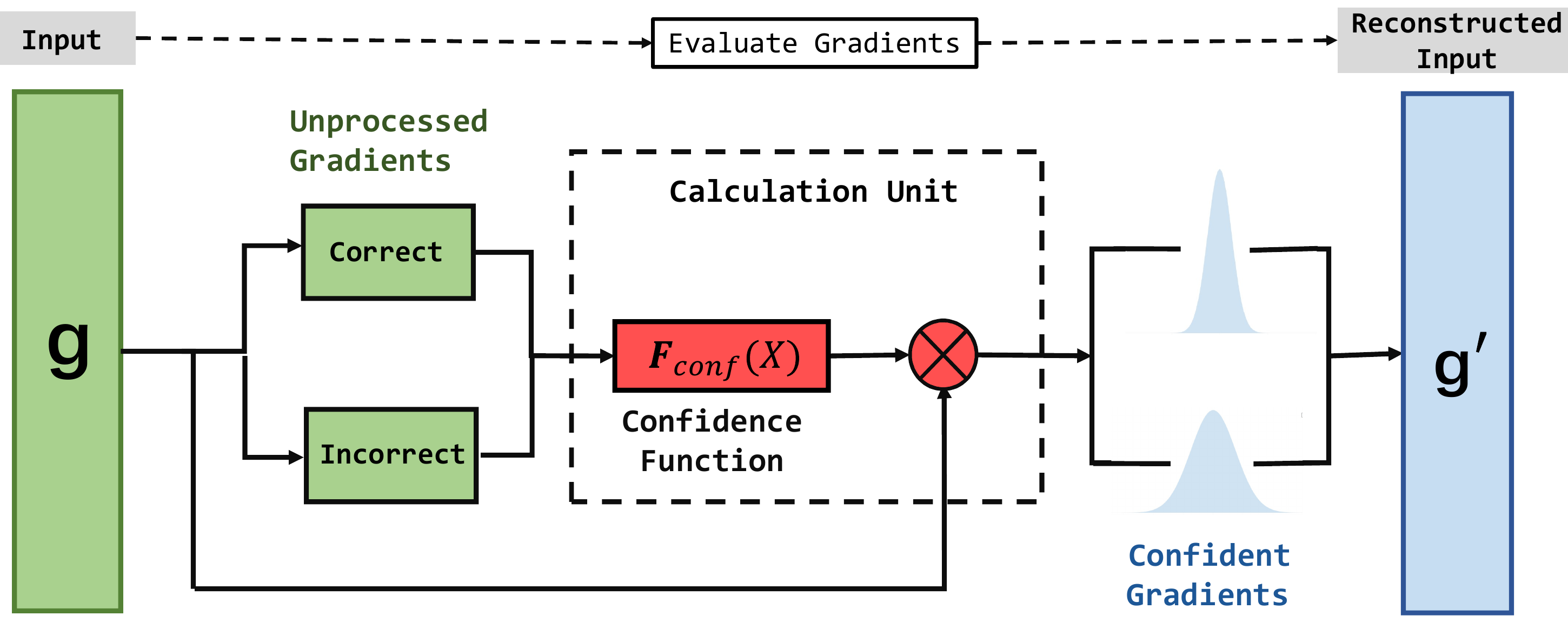}
  \caption{Raw gradients have two parts coming from correct and incorrect samples, respectively. After being processed by the calculation unit, the gradients of correct samples have more zero components than those of the incorrect samples. The input "X" for the confidence function can be data (features), gradients, weights and so on. Because we think there is a close relationship between models, data and learning in deep neural networks. What we know about the optimization in deep learning is not only a non-convex problem, but also related to the model's structure property. In this paper, we show a kind of confidence functions with gradients as its input. So our confidence function is gradient-based.}
\label{process_img}
\end{figure*}

The remainder of this paper is organized as follows. Section 2 discusses some recent advances in optimization algorithms for deep learning. In Section 3, we show our motivation by experimental results and propose a new optimization framework based on the confidence function. In Section 4, we reveal that our signADAM++ can be summarized into the framework. Empirically, our approach consistently achieves significantly better results than other methods on various datasets and models as shown in Section 6.

\section{Related Work}
In this section, we mainly review the recently developed stochastic optimization techniques for deep learning.

\subsection{Traditional Gradient-based Methods}
There are several lines of research studies on the steepest gradient descent method. For training deep neural networks, stochastic gradient descent (SGD) is an efficient method. To solve the problem of navigating ravines \cite{sutton1986two}, the momentum term \cite{qian1999momentum} takes steps in relevant directions frequently. It dampens the oscillation impact along the short axis directions and makes contributions along the long axis directions \cite{rumelhart1985learning}.

Inspired by the DNNs' optimization development and theoretical advance, many adaptive algorithms have been proposed, such as ADAGRAD \cite{duchi2011adaptive}, ADADELTA \cite{zeiler2012adadelta}, ADAM \cite{kingma2014adam}, Nadam \cite{dozat2016Nadam} and AMSgrad \cite{ruder2016overview}. They show great abilities to tackle the issues which have sparse data and non-stationary objectives. These methods attract great attention and they are successfully employed in several applications particularly popular in GANs and Q-learning. However, they also have been observed not to converge in some settings and circumstances. The linkages between adaptive methods especially ADAM and sign-based gradient descent approaches have begun to appear in recent work \cite{balles2017dissecting}, which shows ``sign" is a special case of ADAM under convex assumptions. They think ADAM combines two components: variance adaptation and taking sign, while the latter is a dominant one and they have proven it in greater details.

\subsection{Quantization Methods}
As both the model complexity and the amount of data increases, there are some prior works on quantization of models and optimization to deal with large-scale problems. Inspired by delta-sigma modulation, \citet{seide20141} proposed a compression gradient method in SGD, which uses 1-bit to quantify those gradients during data exchange. This reduces the bandwidth requirement. Surprisingly, quantization can also improve the performance of ADAGRAD in the experiments, which gives us a higher accuracy and shorter training time compared with the common setting. \citet{de2015taming} developed a unified theoretical framework, which has proven that using low-precision arithmetic SGD can converge in a reasonable rate under convex assumptions. Moreover, in \cite{alistarh2017qsgd}, they proposed a method called QSGD, which is applied in the non-convex case and allows a smooth trade-off between convergence and compression per iteration, and they also showed the effectiveness of gradient quantization. Another work \cite{wen2017terngrad} quantified gradients in a different way. It constrains gradients into three values $\{1,0,-1\}$ to speed up distributed training. They proved that gradient clipping is necessary and effective.

The sign-based method was first introduced by \cite{riedmiller1993direct}. They proposed an adaptive method (called RPROP) standing for ``resilient propagation". According to the signs of gradients, RPROP adapts each element update magnitudes. Recently, with the theoretical development of sign-based optimization, \cite{karimi2016linear} and \cite{bernstein2018signsgd} provided the proof of the sign-based method convergence in Polyak-Lojasiewicz condition and non-convex condition, respectively. Moreover, a more recent work \cite{bernstein2018convergence} showed that the sign-based method outperforms existing quantization methods, for it does not need to ensure unbiasedness, while other methods requiring randomization.

\subsection{What We Do}
In this paper, we want to propose an efficient algorithm, which incorporates the sign operation into the adaptive gradient-based methods, e.g., ADAM. At a high level, we propose an efficient framework of optimization algorithms, which evaluate gradients and update models depending on the confidence function setting the gradients to adaptive values in each iteration. This simple approach takes the advantages of a low cost complexity, strong performance guarantee and the adaptive property. For instance, as shown in Section 6, the experimental results reveal our algorithms perform even better than other adaptive and sign-based methods on a variety of both models (e.g., GoogLeNet \cite{szegedy2015going}, VGG-19 \cite{simonyan2014very} and ResNet-18 \cite{he2016deep}) and datasets (e.g., CIFAR-10 and CIFAR-100).

The main contributions of this paper can be summarized as follows:
\begin{enumerate}
  \item We first define a confidence function, which could be thought as augmented sampling and give a common optimization framework for training various DNNs,  as shown in Figure \ref{process_img}.
  \item We then combine the sign operation and the momentum accelerated method with adaptive learning rates into non-convex optimization. The proposed method achieves a satisfactory result, as shown in Section 6.
  \item We provide our algorithms a theoretical guarantee to prove that they at least have the same convergence rate as SIGN-SGD in standard assumptions.
  \item We explain the experimental results from the perspective of the loss landscape and expose the relationship between loss landscape and feature learning.
  \item Experimental results show our motivation is reasonable and it suits the biological neural processes.
\end{enumerate}

\section{MOTIVATION}
Backpropagation (BP) \cite{rumelhart1988learning} is an essential tool to train various DNNs. Using this technique we could improve the accuracy by leveraging information from the training samples. Gradients come from models and data. In this perspective, during updating, gradients are guiding models to extract the unknown features by learning from data. To learn all the features from the training set equally in a better way, we introduce the confidence function and show how it works in the experiment below.

\subsection{How Our Confidence Function Works}
\begin{figure}[H]
\centering
\includegraphics[width=\columnwidth]{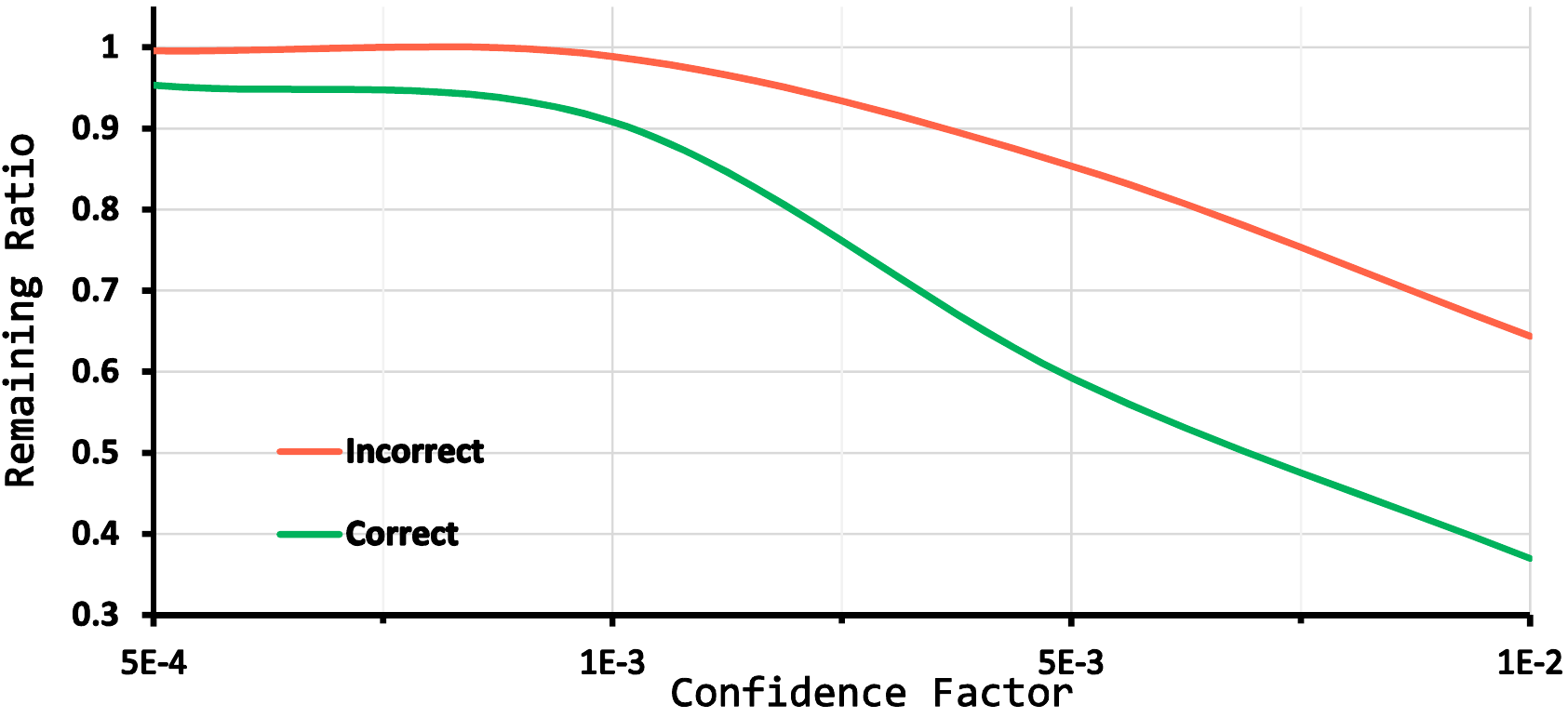}
\vspace{-0.15cm}
\caption{Remaining ratio changes as the confidence factor increases.}
\label{confidence_ratio}
\end{figure}

\textbf{Set-up}\quad We run VGG-Net on CIFAR-10 and randomly choose a layer with many weights as samples. By using different confidence factors, we show the various remaining ratios for correct and incorrect samples, respectively. The definition of the remaining ratio as follows.
\begin{equation*}
\textup{Remaining\;Ratio}=\frac{\|grad\|_2}{\|grad'\|_2},
\end{equation*}
where $grad\in \mathbb{R}^{d}$ is the raw gradient, $grad'$ is the processed gradient, and $\|\cdot\|_2$ denotes the $L_2$-norm.

\textbf{Results}\quad When the confidence factor is set to a proper value, the remaining ratio of correct samples is obviously less than that of the incorrect samples. In addition, we find our confidence function weakens the components of the gradient from both correct and incorrect samples.

\textbf{Discussion}\quad Correct and incorrect samples have different features, which are easy and hard to learn, respectively. And correct samples have more easy features than incorrect samples. Our confidence function has a greater inhibition for the features, which are easy to learn. Also this makes the fitting rate of easy features faster than the hard ones, which lets the models learn the features equally. Of course, our confidence function has a similar effect as online hard sample mining \cite{7553523}. Our improvement is given as follows:
\begin{itemize}
    \item We first implement the augmented sampling in the optimization of deep learning by using our gradient-based confidence function.
    \item Most existing methods only deal with samples, while our method can deal with features. Because hard samples have features easy to learn and easy samples have features hard to learn.
    \item Our method can be more widely used in the deep learning community because it has a great performance on various datasets and models.
\end{itemize}

\begin{table}
\centering
\renewcommand\arraystretch{1.65}
 \resizebox{\columnwidth}{15mm}{
\begin{tabular}{ccc}
\toprule
\multicolumn{1}{l}{} & \multicolumn{1}{c}{Confidence functions}                                                                                                                                                                           & Momentum       \\
\midrule
SIGN-SGD             & $\frac{\delta}{|grad|}$                                                                                                                                                                                           & $\times$       \\
SIGNUM               & ${\delta/|(1-\beta) \sum_{i}^{k}\beta^{t-i}grad_i|}^\ast$                                                                                                                                                     & $\checkmark$   \\
ADAM                 & ${\delta/(1-\beta) \sum_{i=1}^{k}\beta^{k-i}grad_{i}\odot grad_{i}}^\ast$                                                                                                                                     & $\checkmark$   \\
signADAM             & ${\delta/|grad| (1-\beta) \sum_{i=1}^{k}\beta^{k-i}}^{\ast}$                                                                                                                                                                                  & $\checkmark$   \\
signADAM++           & \begin{tabular}[c]{@{}l@{}}$ F_{conf}(grad,\alpha)= \begin{tiny} \begin{cases} 0, & \textup{if} \: grad\in[-\alpha,\alpha], \\ \frac{\delta}{|grad|},& \textup{otherwise} \\ \end{cases} \end{tiny}$ \end{tabular} & $\checkmark$   \\
\bottomrule
\end{tabular}}
\vspace{1mm}
\caption{Comparison of SIGN-SGD, SIGNUM, ADAM, our proposed signADAM and signADAM++ algorithms. Note that ${ }^\ast$ means the above marked form is only approximation to the true confidence function because the algorithms first use the moving average. These methods are not applying the confidence function directly to the unprocessed gradients from the mini-batch samples, which is different from SIGN-SGD and our signADAM++ algorithm.}
\label{framework}
\end{table}

\subsection{Biological Perspective}
 For the view of neural processes, some neuroscience research \cite{douglas2007recurrent} pointed out that cortical neurons are rarely active in their maximum saturation regime. And they suggest that neurons be encoded information in a sparse and distributed way \cite{attwell2001energy}. Additionally, the research revealed that the brain uses a sparse representation: only around $1$-$4\%$ of the neurons are active together at a fixed time \cite{attwell2001energy,lennie2003cost}. When a neuron is not fired, it does not change its state. Similarly, if we regard weights as neurons, neurons not activated represents weights not updated. We can implement this step by assigning small gradients a confidence with zero. In addition, the research from \cite{livnat2008mixability} shows that more independent genes may be the key life keep prosperity in the earth, while in deep learning, learning more distinguished features can make models more robust, which can weaken overfitting.

\subsection{Our Framework Based on The Confidence Function}
We define a new learning framework for training deep neural networks by using our confidence function. It is showed in Algorithm \ref{framework_alg}. We now consider the gradient descent as not only an optimization method but also a learning problem in deep neural networks. Our confidence function makes DNNs know how to learn features and which features to learn. So how to design a simple and efficient confidence function becomes attractive. Because such a new confidence function can help the model learn better features so that the model can get more satisfactory solutions. We also analyze some state-of-the-art optimization methods from the perspective of the confidence function. The analysis is showed in Table \ref{framework}.

\begin{algorithm}
\caption{Generic framework based on our confidence function}
\label{framework_alg}
\begin{algorithmic}
\REQUIRE $\beta$ : Exponential decay rate for moment estimate. \\
\ENSURE$\theta_{0}$ : Initial parameter vector, \\
\qquad \qquad\!\: $\alpha$ : Initial confidence factor. \\
\WHILE{$k \neq $ max epoch}
\STATE $k\leftarrow k + 1$;
\STATE $\tilde{g}_{k}\leftarrow$  Stochastic Gradient$(x_{k-1})$;
\STATE $m_k=\beta   m_{k-1}+F_{conf}(\alpha,\tilde{g}_{k}) \odot  \tilde{g}_{k}$;
\STATE $\theta_{k} \leftarrow \theta_{k-1} - m_k$;
\ENDWHILE
\OUTPUT {$\theta_{k}$.}
\end{algorithmic}
\end{algorithm}

\section{signADAM++}
In ADAM, the real learning rate is inversely proportional to a $L_{2}$-norm of their all current and past gradients. To extend the update rule from $L_{2}$-norm to $L_{\infty}$-norm, then the second moment estimate term becomes a recursive formula: $u_{k}\!=\!\max(\beta_{2}u_{k-1},|g_{k}|)$ in ADAMAX \cite{kingma2014adam}. The update formula will have a sign term when $\beta_{2}  u_{k-1} \leqslant |g_{k}| $. Furthermore, the gradient descent using the sign can be viewed as the steepest descent with $L_{\infty}$-norm \cite{boyd2004convex}. \citet{balles2017dissecting} proved that the ADAM's update rule is dominated by taking the sign of stochastic gradients. In addition, by setting the hyper-parameters $\beta_{1}$ and $\beta_{2}$ in ADAM to zero ($\beta_{1}, \beta_{2}\rightarrow 0$), ADAM can be changed into SIGN-SGD \cite{bernstein2018signsgd}.

In the optimization field, to combine the advantages of ADAM and SIGN-SGD, we first design an algorithm, called signADAM. Another idea of this algorithm is to reinforce the sign aspect in ADAM. Meanwhile, using ``sign" can make communication and calculation effective. ADAM can be easily incorporated with ``sign" merely by taking the ``sign" of all stochastic gradients in ADAM.

\begin{algorithm}
\caption{signADAM}
\label{alg1}
\begin{algorithmic}
\REQUIRE $\delta$ : Learning rate.\\
$\beta_{1},\beta_{2} \in [0,1)$ : Exponential decay rates for moment estimates, $\epsilon \leftarrow$ 1e-8.\\
$\theta_{0}$ : Initial parameter vector.\\
$m_{0} \leftarrow 0$, $v_{0}\leftarrow 0$, $k \leftarrow 0$.\\ Current point $x_{k}$, current first and second moment $m_{k},v_{k}$.
\WHILE{$k \neq $ max epoch}
\STATE $k\leftarrow k + 1$;
\STATE $\tilde{g}_{k}\leftarrow $ Stochastic Gradient$(x_{k-1})$;
\STATE $ \tilde{g}_{k} \leftarrow sign(\tilde{g}_{k})$;
\STATE $m_{k}\leftarrow \beta_{1}   m_{k-1}+(1-\beta_{1})  \tilde{g}_{k}$; //(update first moment estimate)
\STATE $v_{k}\leftarrow \beta_{2}   v_{k-1}+(1-\beta_{2})  \tilde{g}_{k}\odot \tilde{g}_{k} $;  //(update second moment estimate)
\STATE $\theta_{k}\leftarrow \theta_{k-1}- \delta   \frac{m_{k}}{\sqrt{v_{k}}+\epsilon}$; //(update parameters)
\ENDWHILE
\OUTPUT {$\theta_{k}$.}
\end{algorithmic}
\end{algorithm}

For each dimension of the gradients, the second moment estimate term in our signADAM algorithm is a value determined by the iteration $k$. Since $[sign(\tilde{g}_{k})]^{2}$ is a fixed number equal to 1, so the term $v_{k}$ can be rewritten as follows:
\begin{equation*}
v_{k}= -\beta_{2}^{k-1}+1.
\end{equation*}
As the global iteration $k$ increases, the magnitude of the difference between 1 and $v_{k}$ would become infinitesimal ($k\rightarrow \infty, v_{k}\rightarrow 1$).
In numerical the effect of this term can be neglected. Thus the new parameter update rule is formulated as follows:
\begin{equation*}
\theta_{k} \leftarrow \theta_{k-1}-\delta   m _{k}.
\end{equation*}

Then we analyze ADAM and SIGN-SGD from the perspective of the confidence function.
ADAM gives an approximate adaptive confidence to all gradients. In contrast, SIGN-SGD sets the confidence to 0 only when the gradients equal to 0 and gives other gradients a real adaptive confidence. \citet{bernstein2018signsgd} pointed out that SIGN-SGD belongs to the family of the adaptive methods. This confirms our adaptive confidence on large gradients for both ADAM and SIGN-SGD. We have shown the confidence functions of all the algorithms in Table \ref{framework}.

If we set an appropriate confidence factor in our confidence function (as shown in Figure \ref{confidence_ratio}), there will be different effects on the unprocessed gradients from both correct and incorrect samples. The confidence factor can be used to control the sparsity of the gradients. Considering the motivations in Section 3 and the framework based on our confidence function, we design a new confidence function and apply it into signADAM. The new algorithm is called signADAM++.

We use the following confidence function in the proposed signADAM++ algorithm:
$$ F_{conf}(grad,\alpha)=
 \begin{cases}
   0, & \textup{if} \: grad\in[-\alpha,\alpha], \\
   \frac{\delta}{|grad|},& \textup{otherwise}, \\
 \end{cases}
$$
where $\alpha$ is the confidence factor.

From a computational point of view, choosing the sign operation is appealing for the following reasons.
\begin{itemize}
  \item Tasks based on deep learning such as image classification usually have a large amount of training data (e.g., the popular image dataset, ImageNet, has 1.5 millions images). ``Training" on such dataset may take a week even a month to shrink calculation in every step which can enhance the efficiency.
  \item The algorithm can induce sparsity of gradients by not updating some gradients. signADAM++ makes gradients sparse and leads models to learn distinguished information. \citet{reddi2018convergence} indicated that it is important to update large and informative gradients. One of the ADAM's drawbacks is to ignore these gradients by using the exponential average algorithm \cite{kingma2014adam}. Our algorithm can prevent this issue by giving small gradients a very low confidence (e.g., 0).
  \item Nowadays, learning from samples in deep learning is always stochastic due to the giant amount of data and models. In fact, the confidence of some gradients produced by loss functions and samples should be low. From the perspective of maximum entropy theory, each feature should have the equal right to make efforts in deep neural networks. This method will make the models work well.
  \item Although there are large gradients caused by the incorrect samples, large gradients may hurt models' generalization and learning ability found by \cite{luo2019adaptive} and \cite{reddi2018convergence}. signADAM++ addresses this issue by using moving average after applying confidence for unprocessed gradients and an adaptive confidence for some large gradients. These can give models a better performance.
\end{itemize}

\begin{algorithm}
\caption{signADAM++}
\label{alg2}
\begin{algorithmic}
\REQUIRE $\delta$ : Learning rate.\\
\qquad \;$\beta \in [0,1)$ : Exponential decay rates for moment estimates.\\
\qquad \;$\alpha$ : Confidence factor.\\
\ENSURE $\theta_{0}$ : Initial parameter vector, $m_{0} \leftarrow 0$ (Initial moment vector), $k \leftarrow 0$  (Initial global step).
\\Current point $x_{k}$, and current momentum $m_{k}$.
\WHILE{$k \neq $max epoch}
\STATE $k\leftarrow k + 1$;
\STATE $\tilde{g}_{k}\leftarrow $ Stochastic Gradient$(x_{k-1})$;
\IF{$ \left | \tilde{g}_{k} \right | < \alpha$}
\STATE $ \tilde{g}_{k} \leftarrow 0$;
\ELSE
\STATE $ \tilde{g}_{k} \leftarrow sign(\tilde{g}_{k})$;
\ENDIF
\STATE $m_{k}\leftarrow \beta   m_{k-1}+(1-\beta)  \tilde{g}_{k}$; //(update first moment estimate)
\STATE $\theta_{k}\leftarrow \theta_{k-1}- \delta   m_{k}$; //(update parameters)
\ENDWHILE
\OUTPUT{ $\theta_{k}$.}
\end{algorithmic}
\end{algorithm}

\section{CONVERGENCE ANALYSIS}
In the training process of deep neural networks, given a finite length sequence of input and output pairs, the goal is to minimize the following empirical loss,
\begin{equation}
f(\theta):= \frac{1}{N}\sum_{i=1}^{N}f_{i}(\theta),
\label{eq:lossfunction}
\end{equation}
where the loss function $f_{i}(\theta)$ is discrepancy between the model output and ground truth. A neural network can be regarded as a function with the parameter $\theta\in \mathbb{R}^{d}$ and produces an output for each input $\xi^{i}$. We aim to analyze the objective function $f(\theta,\xi^{i})$, which is a non-convex function of its parameters. Thus the optimization problem is given by
\begin{equation}
\theta^{\ast} = \arg\min_{\theta\in \mathbb{R}^{d}}\, f(\theta).
\label{eq:optpro}
\end{equation}
To simplify the problem, we make several assumptions. It would not loss anything compared to typical SGD assumptions, since these assumptions are obtained from SGD.

\begin{assumption}[Global minimal]
For all parameters $\theta \in \mathbb{R}^{d}$, there exists a $d$-dimensional vector $\theta^{\ast}$ that satisfies the following inequality
\begin{equation}
f(\theta)\geqslant f(\theta^{\ast}).
\end{equation}
\end{assumption}

\begin{assumption}[$L$-smooth]
Let $g(x)$ denote the real gradient of the objective function $f(\cdot)$ evaluated at point x. Then $\forall x,y$, we require that for some non-negative constants $\vec{L}:=[L_1,\cdots,L_d]$,
\begin{equation}
\left | f(y)-[f(x)+g(x)^T(y-x)] \right | \leqslant \frac{1}{2}\sum_{i}^{ }L_i(y_i-x_i)^2.
\end{equation}
For the twice differentiable function $f(\cdot)$, this implies that $-diag(\vec{L}) \prec H \prec diag(\vec{L})$. This is related to the slightly weaker coordinate-wise Lipschitz condition used in the block coordinate descent literature\cite{richtarik2014iteration}.
\end{assumption}

\begin{assumption}[Variance bound]
\noindent Since  $x \in \mathbb{R}^{d}$ is stochastic, its homologous gradient $\tilde{g}$, which has coordinate bounded variance, is an independent unbiased estimate for the real gradient ${g}$.
\begin{equation}
\mathbb{E}[\tilde{g}(x)]=g(x),\;\mathbb{E}[(\tilde{g}(x)-g(x))^{2}]\leqslant {\sigma_{i}}^{2},
\end{equation}
for a vector containing non-negative constants $\Vec{\sigma}:=[\sigma_{1},\cdots ,\sigma_{d}]$
\end{assumption}

 Assumptions 2 and 3 are the standard assumptions of Lipschitz smoothness and total bounded variance, respectively. Under these assumptions and the framework of SIGN-SGD \cite{bernstein2018signsgd}, we have the following results.

\begin{theorem}[Non-Convex Convergence Rate]
\label{theo1}
Suppose Assumptions 1, 2 and 3 hold. Let $\{\tilde{g}_k\}$ ($k=0, \cdots,K$) be a sequence generated by Algorithm 1, where the learning rate is set to $\delta _k:=1/\sqrt{\| \overrightarrow{L}\|_{1} K }$ and the size of mini-batch as $n_k=K$.
Let N be the cumulative number of stochastic gradient calls up to step $K$ (i.e.,$ N = O(K^{2})$). Then we have
\begin{equation*}
\begin{split}
&\;\mathbb{E}\!\left[\frac{1}{K}\sum_{k=0}^{K-1}\left \| g_{k} \right \|_{1}\right]^{2} \\
\leqslant&\; \frac{1}{\sqrt{N}}  \left[\sqrt{\left \| \vec{L} \right \|_{1}}\left(f(\theta_{0})-f(\theta^{\ast})+\frac{1}{2}+\frac{8\beta}{(1\!-\!\beta)^2}\right)+2\left \| \vec{\sigma} \right \|_{1}\right]^{2} \!.
\end{split}
\end{equation*}
\end{theorem}

To prove Theorem 1, we first present and prove the following lemmas. In the following lemma, we apply mathematical induction for exponential average of the gradients' signs.

\begin{lemma}
For all $ m_{k} \in \mathbb{R}^{d}$, they satisfy the following inequality
\begin{equation}
\left | m_{k,i} \right | \leqslant{1},\;\; i\in \{1,2, \cdots,d\},
\end{equation}
where $\left | m_{k,i} \right |$ is the exponential average of the gradients' sign.
\end{lemma}
The detailed proofs of Theorem 1, Lemma 1 and the lemma below are provided in the APPENDIX. Next we give a lemma, which shows the exponential average of gradients' signs.

\begin{lemma}
$\frac{1-\beta}{1-\beta^k}\sum^{K-1}_{t=0}\beta^t[||g_{k-t} - g_{k}||_1]\leq 4\frac{\sqrt{\left \| \vec{L} \right \|_{1}}}{\sqrt{K}}\frac{\beta}{(1-\beta)^{2}}$.
\end{lemma}

\section{EXPERIMENTS}
In this section, we discuss our empirical evaluation of signADAM++ and other related algorithms such as ADAM \cite{kingma2014adam}, SIGN-SGD and SIGNUM \cite{bernstein2018signsgd}. We study the classical deep learning problems (e.g., image classification) and conduct several experiments for different state-of-the-art DNNs such as GoogLeNet \cite{szegedy2015going}, VGG-19 \cite{simonyan2014very}, ResNet-18 \cite{he2016deep} and SeNet-18 \cite{hu2018squeeze}. Using the torch-rnn library, we also train and test an LSTM network.

\begin{figure}[!th]
\centering
\subfigure[VGG19: Log Loss]{\includegraphics[width=0.4938\columnwidth]{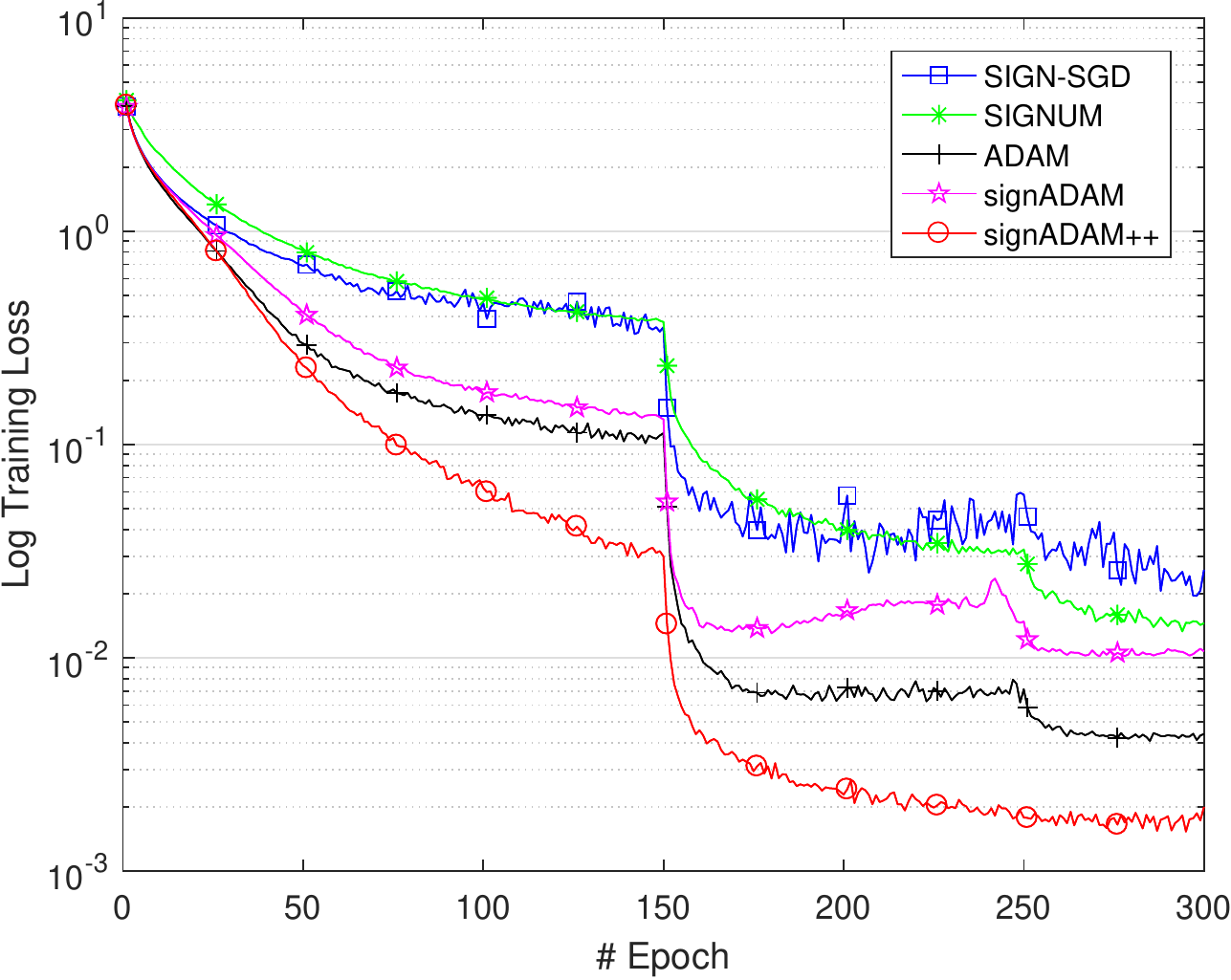}}
\subfigure[VGG19: Log Test Error ]{\includegraphics[width=0.4936\columnwidth]{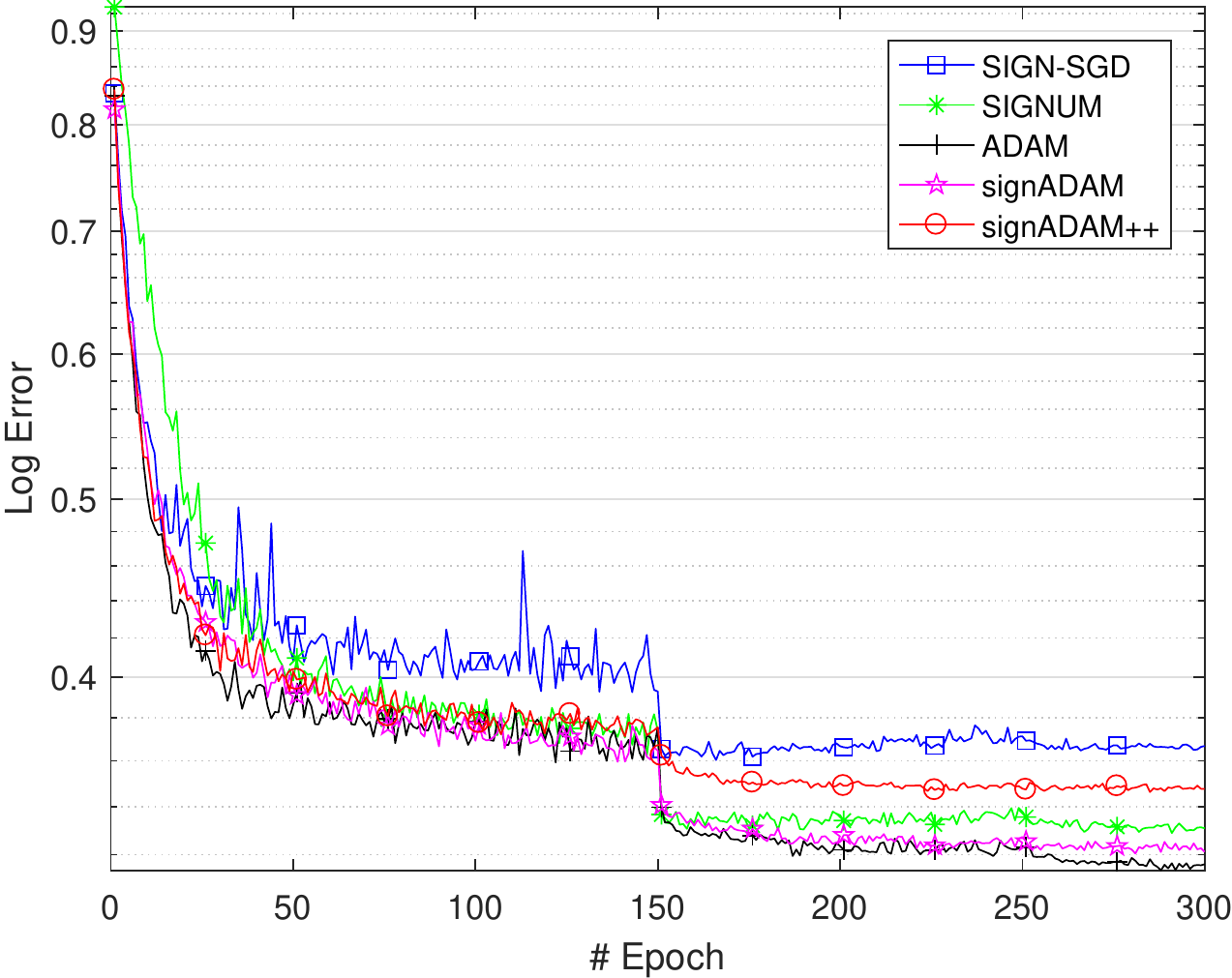}}
\vspace{-4mm}
\caption{Comparison of the training loss and test error of all the algorithms on CIFAR-100.}
\label{cifar100}
\end{figure}

\subsection{Setup}
We compare the proposed algorithms (i.e., signADAM and signADAM++) with SIGN-SGD \cite{bernstein2018signsgd}, SIGNUM \cite{bernstein2018signsgd} and ADAM \cite{kingma2014adam}. We run all the experiments on the three standard image datasets: MNIST, CIFAR-10 and CIFAR-100, all of which are divided into two parts: the training set and test set.

In the experiments on CIFAR-10 and CIFAR-100, all the training images are resized to $32\times32\times3$. These images are then randomly cropped to $32\times32\times3$, with padding equals to 4. They are also randomly flipped and finally normalized. For the test images, they are only normalized and no other data augmentation tricks are used in the training and test processes. In the experiments on MNIST, the training and test images are both normalized. We record the objective loss on the training set and test error on the test set and evaluated algorithms in the following deep learning models.

\begin{itemize}
\item   LeNet consists of two sets of convolutional and average pooling layers, followed by a flattening layer, then two fully-connected layers and finally a softmax classifier.
\item   VGG-19 uses 16 convolution layers and 3 fully connected layers.
\item   GoogLeNet consists of 22 layers,  using the Inception module to make itself deeper than VGG.
\item   ResNet-18 increases depth of the network results in less extra parameters and reduces the effect of gradient vanishing problems. ResNets achieve the improvement by adding simple skip connections.
\item   SENet-18 is an architectural using dynamic channel-wise feature recalibration, which can improve the representational power of a network.
\item   Two layers of LSTM units for the classification of MNIST.
\end{itemize}

We first test LeNet and LSTM on the MNIST dataset, which can be viewed as shallow neural networks. Then to verify our algorithms for training DNNs, we evaluate the performance of all the algorithms on the stat-of-the-art DNNs (e.g., VGG-19, GoogLeNet, ResNet-18 and SeNet-18). In addition, to prove that our algorithms work well on different datasets, we also design experiments on the CIFAR-100 dataset by running VGG-19. For all experiments, we have used optimal hyperparameters for all algorithms and applied learning rate schedulers for all algorithms. We also implement the $L_2$ normalization proposed by \cite{loshchilov2017fixing} for all methods. And the hyperparameters are given as follows. In experiments on MNIST, we use 0.001 as the learning rate of SIGN-SGD, SIGNUM and ADAM. On CIFAR-10 and CIFAR-100, we use 0.0001 as their learning rate. In all the experiments, we use the recommended initial values for $\beta_1$ and $\beta_2$ (i.e., $\beta_1\!=\!0.9$, $\beta_2\!=\!0.999$) and set the regularization coefficient to 5e-4.

\begin{figure}[h]
\centering
\includegraphics[width=\columnwidth]{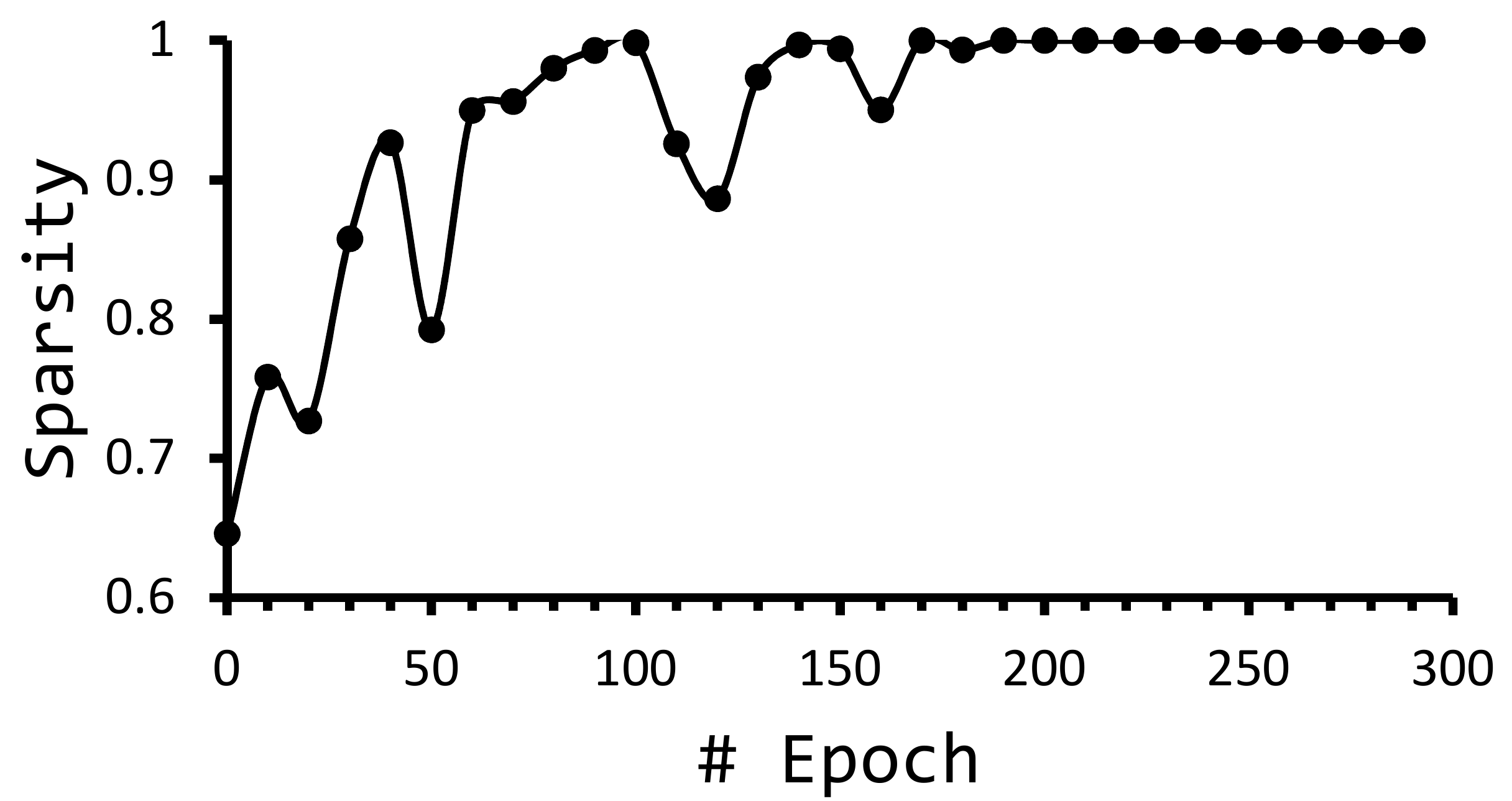}
\vspace{-4mm}

\caption{Sparsity of confidence gradients.}
\label{sparse}
\end{figure}

\begin{figure*}[!th]
\centering
\subfigure[LeNet: Log Loss]{
\includegraphics[width=0.499\columnwidth]{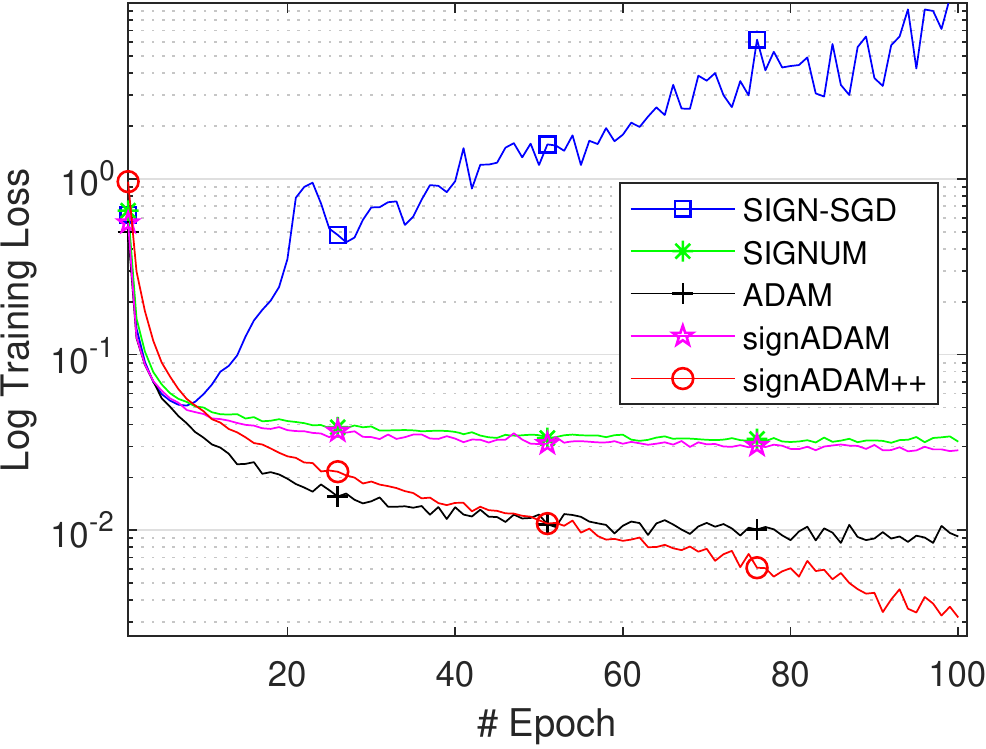}}
\subfigure[LeNet: Log Test Error]{
\includegraphics[width=0.499\columnwidth]{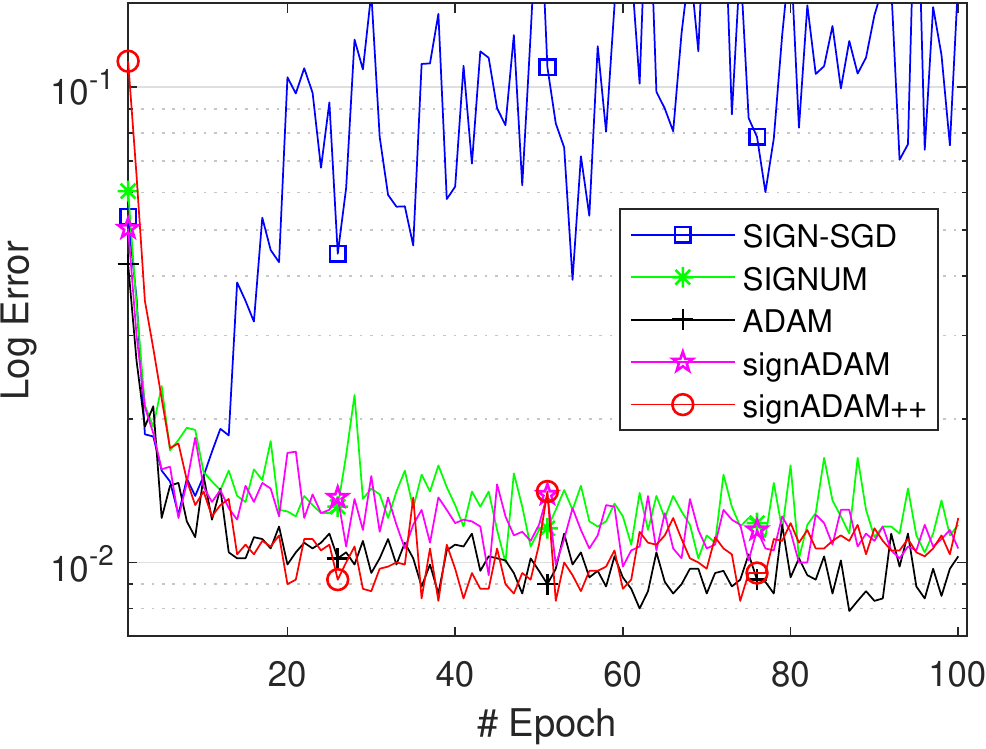}}\;
\subfigure[LSTM: Log Loss]{
\includegraphics[width=0.499\columnwidth]{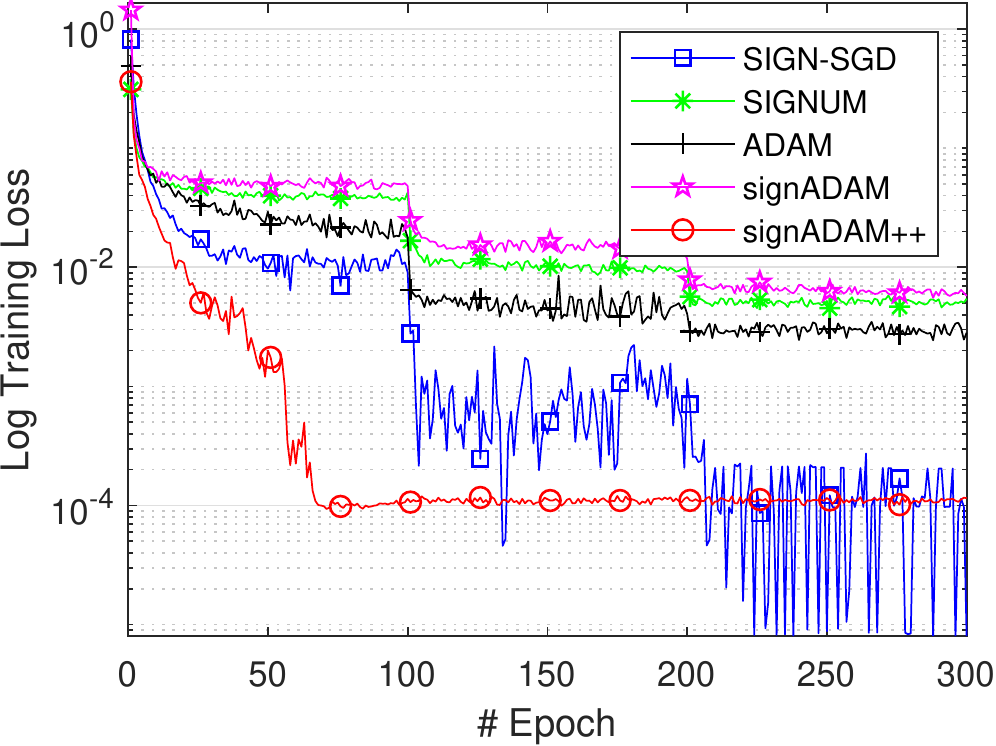}}
\subfigure[LSTM: Log Test Error]{
\includegraphics[width=0.499\columnwidth]{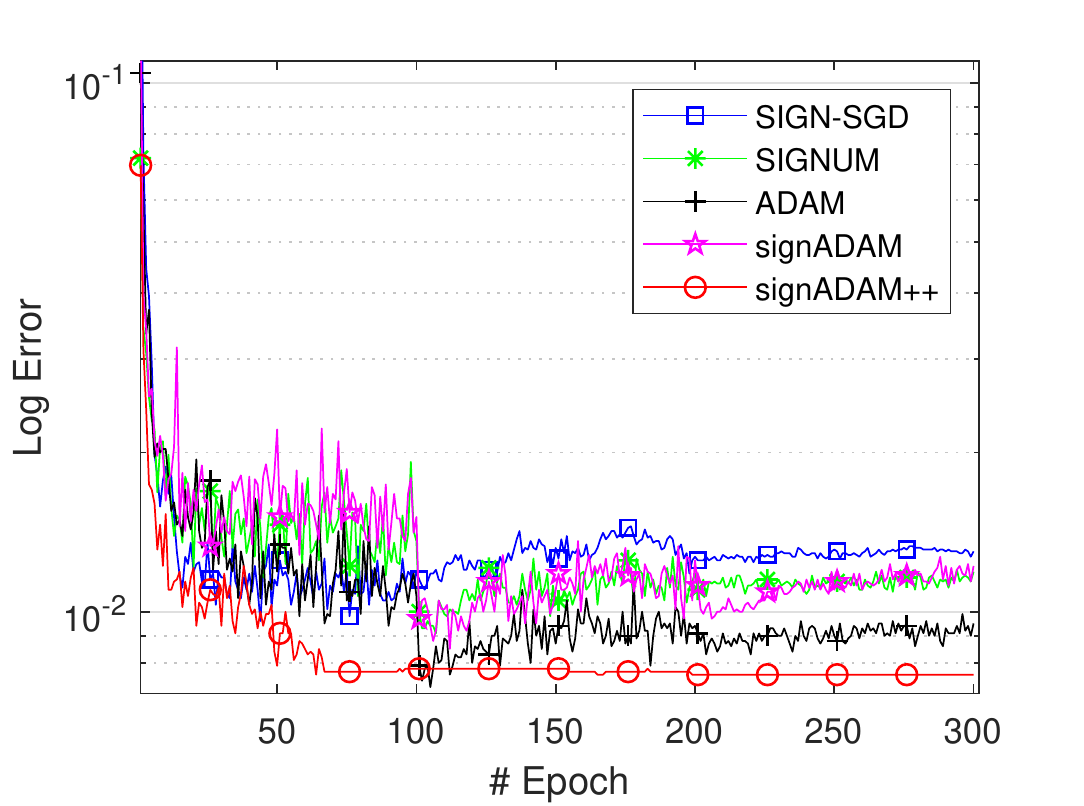}}
\vspace{-4mm}

\caption{Comparison of the logarithm training loss and test error of all the algorithms on MNIST.}
\label{MNIST}
\end{figure*}

\begin{figure*}[!th]
\centering
\subfigure[VGG19: Log Loss]{
\includegraphics[width=0.499\columnwidth]{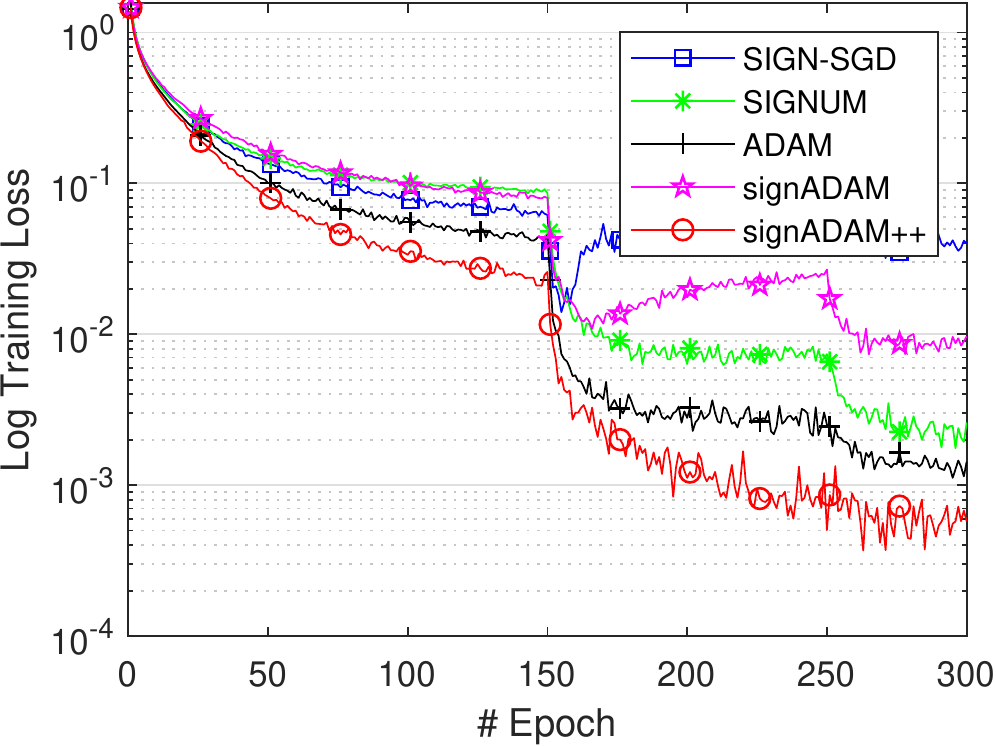}}
\subfigure[VGG19: Log Test Error]{
\includegraphics[width=0.499\columnwidth]{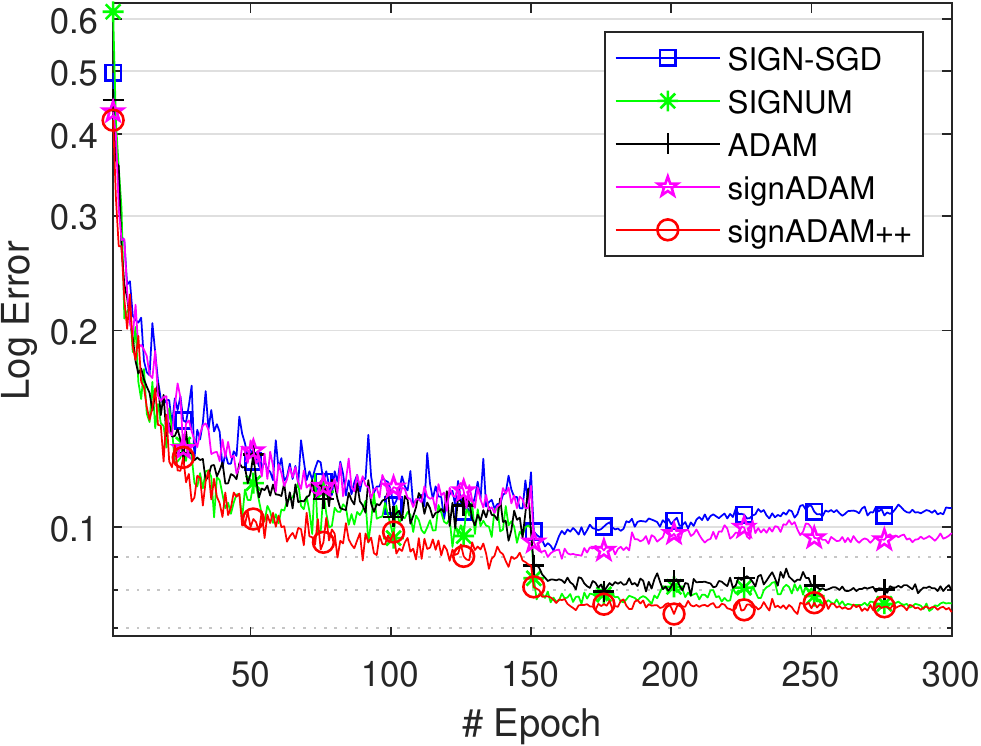}}\;
\subfigure[GoogLeNet: Log Loss]{
\includegraphics[width=0.499\columnwidth]{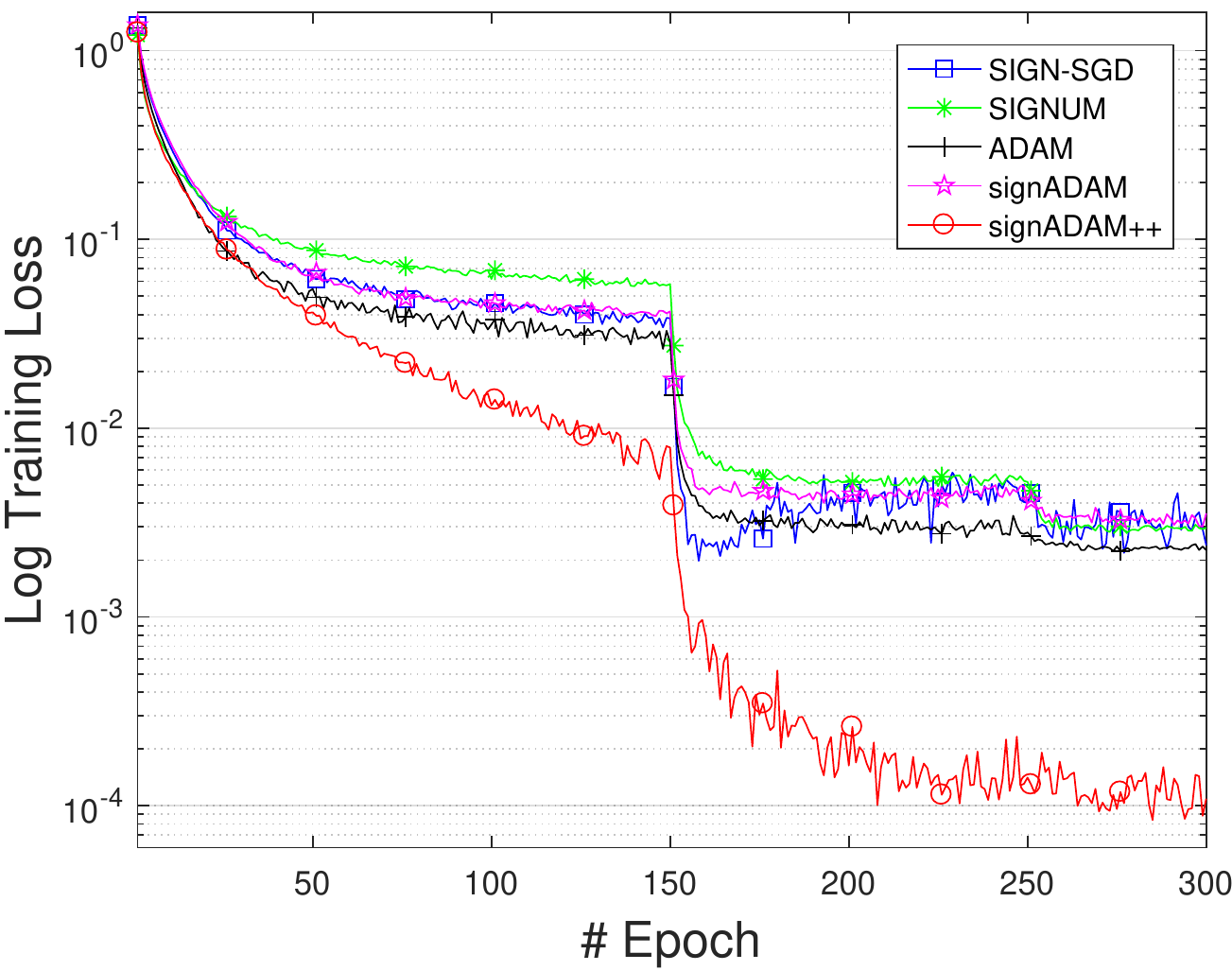}}
\subfigure[GoogLeNet: Log Test Error]{
\includegraphics[width=0.499\columnwidth]{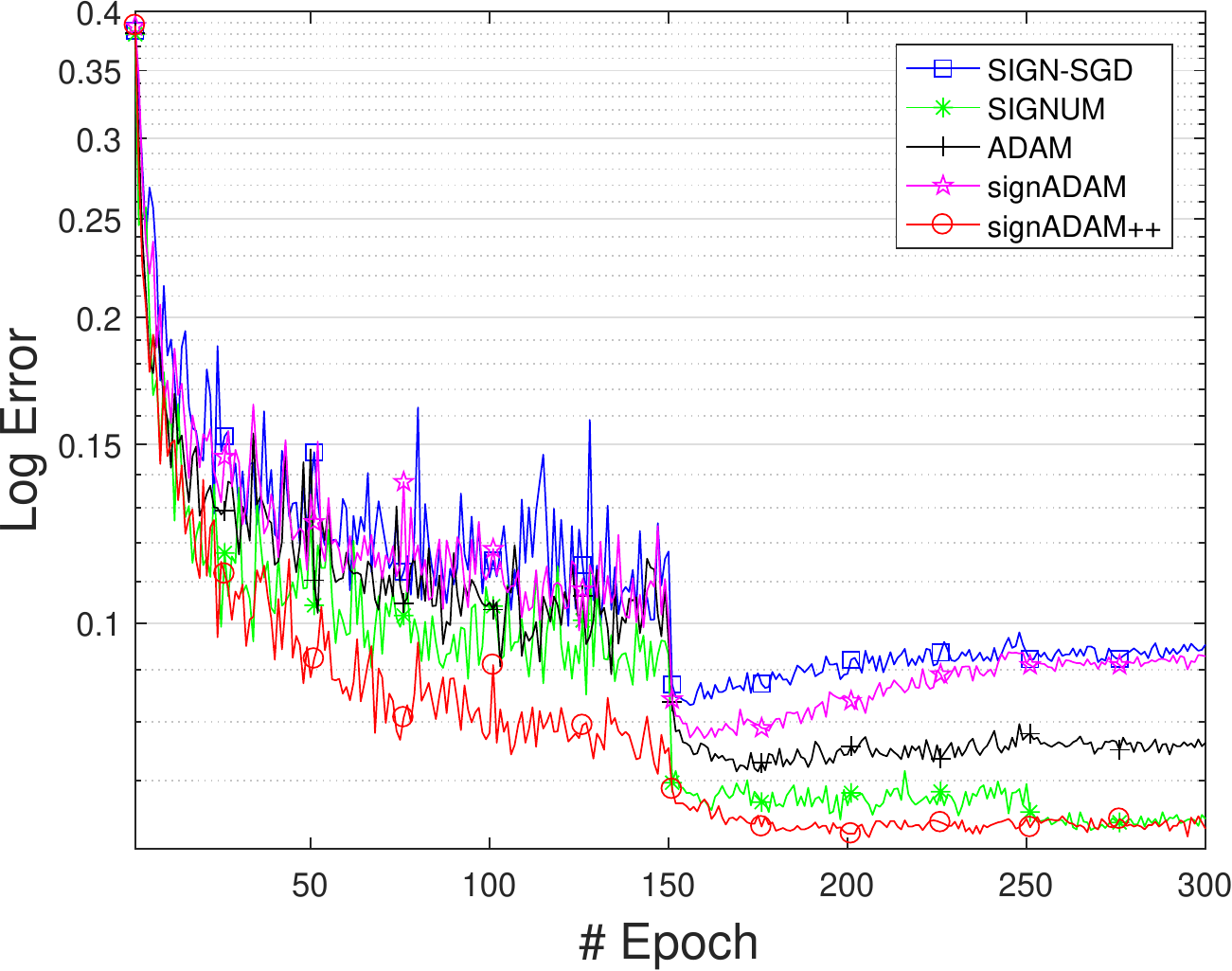}}

\subfigure[ResNet18: Log Loss]{
\includegraphics[width=0.499\columnwidth]{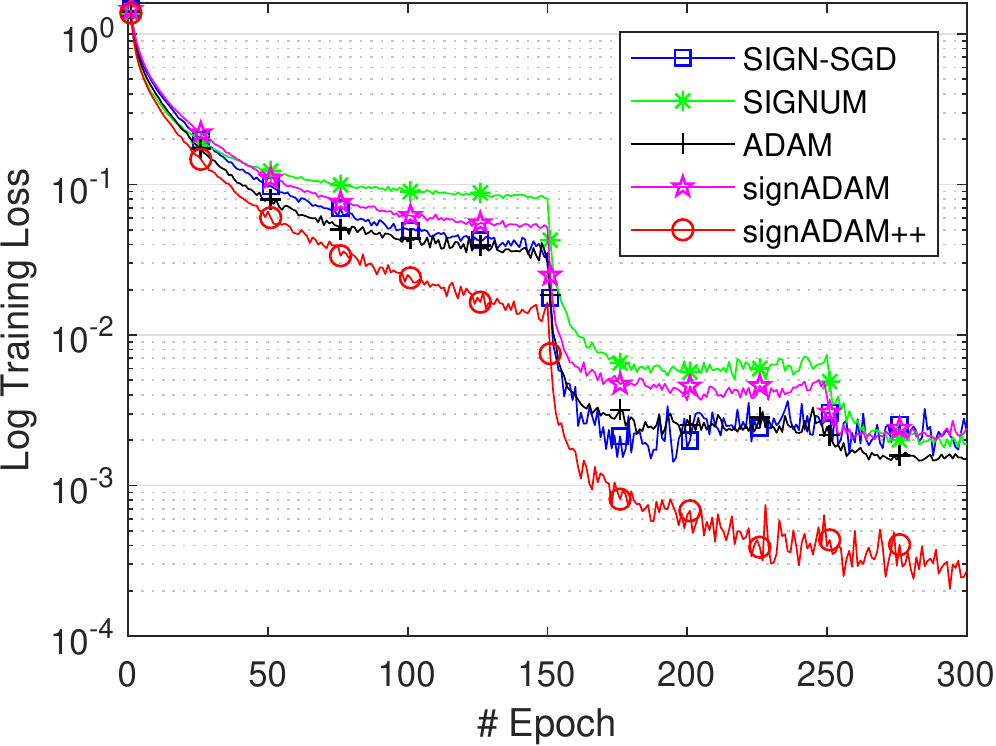}}
\subfigure[ResNet18: Log Test Error]{
\includegraphics[width=0.499\columnwidth]{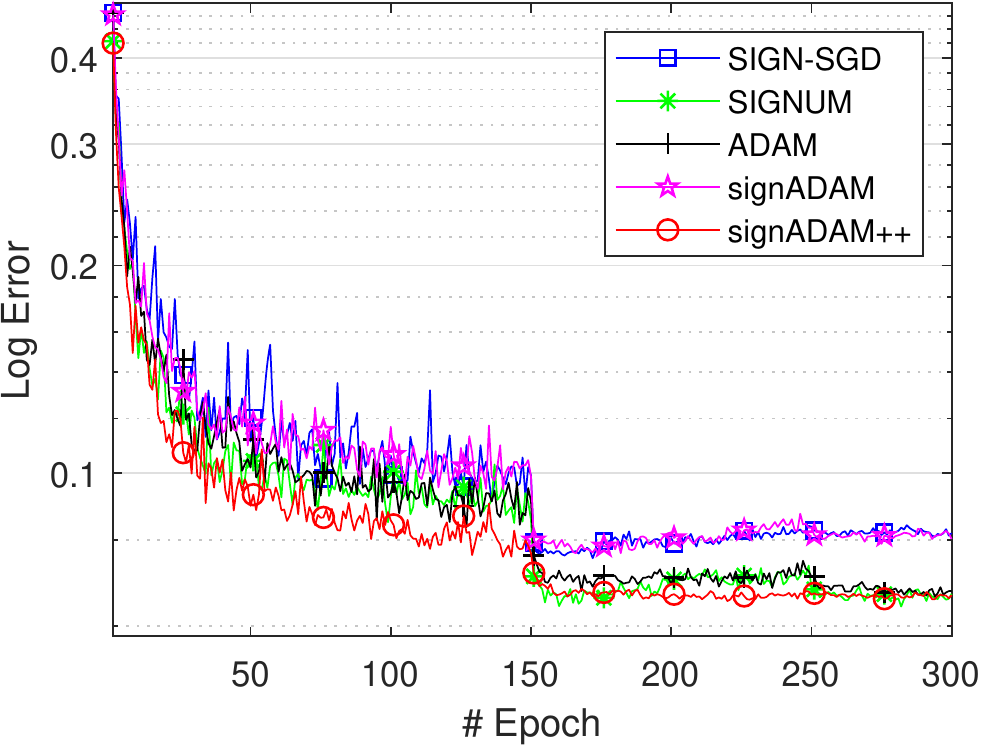}}\;
\subfigure[SENet18: Log Loss]{
\includegraphics[width=0.499\columnwidth]{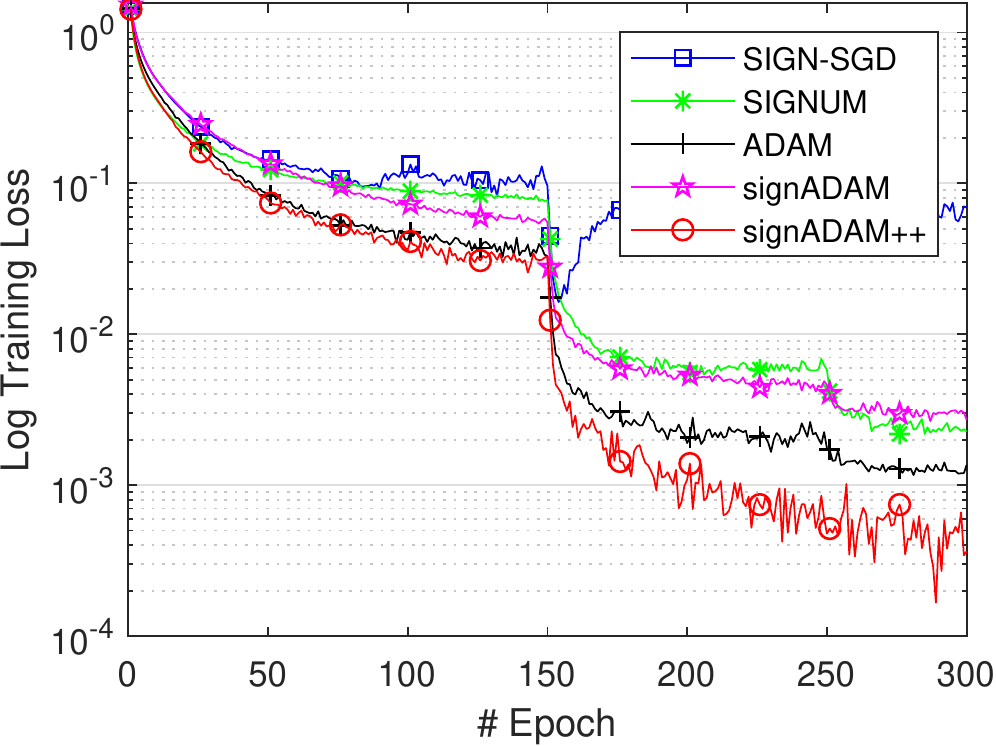}}
\subfigure[SENet18: Log Test Error]{
\includegraphics[width=0.499\columnwidth]{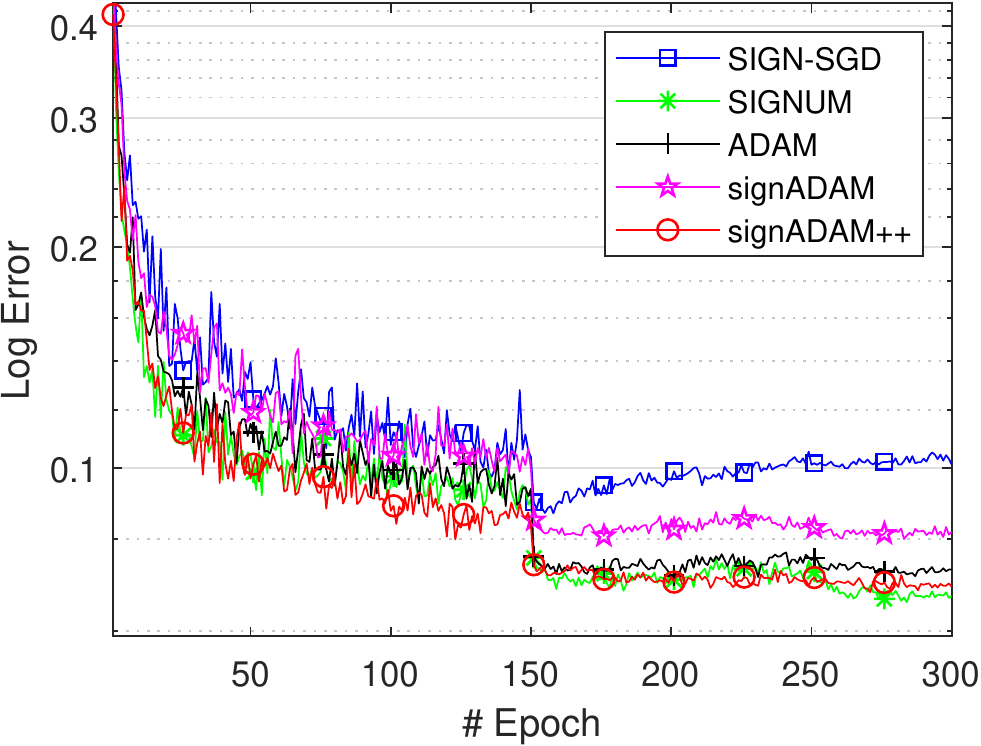}}
\vspace{-4mm}

\caption{Comparison of the logarithm training loss and test error of all the algorithms on CIFAR-10.}
\label{cifar10}
\end{figure*}

\subsection{Experimental Results}
Figures \ref{cifar100}, \ref{MNIST} and \ref{cifar10} show the experimental results of all the algorithms. From all the results, we make the following observations.

\subsubsection{\textbf{Sparsity}}
\paragraph{\textbf{Figure \ref{sparse} shows the sparsity of the confidence gradients}} These parameters are sampled from a randomly chosen layer of VGG-19 on the CIFAR-10 dataset. We find that the gradients are indeed sparse when applying our confidence function into the unprocessed gradients. It fits the biological neural process as mentioned in Section 3. By observing the training loss on CIFAR-10, our signADAM++ enjoys faster convergence than the other state-of-the-art algorithms. Besides, the effect of our confidence factor is similar as the the hard threshold methods in \cite{donoho1994ideal}, which is used to tackle with Wavelet noises. The effectiveness of sparsity can be also confirmed by ReLU activations, as shown in \cite{nair2010rectified,glorot2011deep,langford2009sparse}.

Our method is to regard optimization as a better learning strategy for models. We decrease the randomness created by the stochastic mini-batch, which can increase the convergence rate. As to the feature learning, we tend to make the models learn all the features equally \cite{tishby2015deep}.

\subsubsection{\textbf{Training Loss}}
\paragraph{\textbf{signADAM++ has the fastest convergence speed}} The training loss of signADAM++ is distinguished from that of SIGN-SGD, SIGNUM, ADAM and signADAM during the entire experiments. signADAM++ makes rapid initial progress, and achieves lower loss than other algorithms. There is a big reduction at epoch 150 after we apply learning-rate-decay to all the algorithms. It means that signADAM++ can reach our expectation in a shorter time. We infer that our signADAM++ has removed some randomness so that the parameters can be updated in main and correct directions. This can speed up the training of deep neural networks. In other words, thanks to our reasonable confidence function, the significant samples and their corresponding features have been learned by neural networks preferentially.

\paragraph{\textbf{signADAM performs like SIGNUM}}
We notice that the curves of signADAM and SIGNUM are close especially in the experiments on CIFAR-10 and MNIST. There are two following main reasons:
\begin{itemize}
    \item signADAM and SIGNUM both have moving average and sign operation. Although moving average indeed can help accelerate in the relevant directions and dampen oscillations, some small oscillations cannot be completely removed because its momentum coefficient is not adaptive. These small
    oscillations are increased by the sign operation. Similarly, signADAM strengthens the small oscillations by using the sign operation and dampens oscillations by using the moving average.
    \item Considering that the high precision of the gradients, the square of the sign for a gradient is close to a constant as the iteration increases. So we can merge the second order estimate into the learning rate. Our experimental results indeed confirm this statement.
\end{itemize}

\paragraph{\textbf{SIGN-SGD has large oscillations}} SIGN-SGD updates parameters by using the signs of its gradient. This means that SIGN-SGD does not distinguish the components of gradients. It makes neural networks learn some samples and features so fast that SIGN-SGD finds a local minimal with higher loss.

\subsubsection{\textbf{Test Error}}
\paragraph{\textbf{signADAM++ has a competitive optimization rate.}}
On the test set, we find that signADAM++ also has a satisfactory convergence rate. Table 2 shows the numbers of epochs for all the algorithms when the error reaches a given tolerance. In the experiments of MNIST and CIFAR-10, signADAM++ performs well. However, signADAM++ performs slightly worse on CIFAR-100. We give the following explanation: Because there are 100 classes in CIFAR-100, each class in CIFAR-100 has obviously fewer samples than that in CIFAR-10. So we do not have enough samples for each class to learn. Thanks to the adaptive property of ADAM and SIGNUM, they can both continue learning some features from samples regardless the confidence of information. But for signADAM++, because we apply a very low confidence (e.g., 0) into some information, this makes some features hard even impossible to be learned so specific as CIFAR-10. In other words, due to the lack of data, features on CIFAR-100 are not treated as equally as on CIFAR-10. This result in that the optimal point is located around a sharp minimal, which is usually considered to have a worse generalization than a flat minimal. The adaptive methods can have a large step size in some directions to skip out of the sharp minimal points. And it learns some features relatively sufficient. So the confidence factor decides which features to learn. To keep the models learning features from samples all the time and finding some flatter minimal points, we develop an adaptive method to solve the problem in the next subsection.

\begin{table*}[!th]
\setlength{\tabcolsep}{4mm}\begin{tabular}{c|c|c|c|c|c|c|c}
\hline
           & \multicolumn{2}{c|}{MNIST (0.01)}        & \multicolumn{4}{c|}{CIFAR-10 (0.1)}                                                         & CIFAR-100 (0.4) \\
            \hline
           & LeNet       & \multicolumn{1}{l|}{LSTM} & VGG-19      & GoogLeNet   & \multicolumn{1}{l|}{RseNet-18} & \multicolumn{1}{l|}{SENet-18} & VGG-19         \\ \hline
SIGN-SGD   & -           & 44                        & 151         & 31          & 49                             & 83                            & 91             \\ %\hline
SIGNUM     & 46          & 101                       & 72          & 39          & 76                             & 71                            & 48             \\ %\hline
ADAM       & \textbf{20} & 58                        & 123         & \textbf{23} & 41                             & 45                            & \textbf{31}    \\ %\hline
signADAM   & 44          & 101                       & 151         & 32           & 57                             & 69                            & 39             \\ %\hline
signADAM++ & \textbf{20} & \textbf{21}               & \textbf{50} & 24          & \textbf{37}                    & \textbf{41}                   & 46             \\
 \hline
\end{tabular}
\vspace{1mm}
\caption{The number of epochs for all algorithms to reach a constant error. For MNIST, the error tolerance is 0.01, while it is 0.1 and 0.4 for CIFAR-10 and CIFAR-100, respectively. Note that ``-" means that the algorithm does not achieve the expected error in the entire run.}
\end{table*}

\subsection{Our Adaptive and Weight-decay Technique}
Considering that the ``stop learning" issue above, we improve our algorithm and introduce the adaptive confidence function. In particular, we set the confidence factor in the confidence function to a dynamic value, which is determined by the standard deviation of the gradients. So the new confidence function is defined as follows.
 \begin{align*}
  m_t &=  \beta_1 m_{t-1}+(1-\beta_1)\sqrt{var(grad)},\\
  \alpha_t &=\beta_2 m_t,
 \end{align*}
 where $\beta_1$ and $\beta_2$ are the hyper-parameters in the moving average, and $\alpha$ is the confidence factor. By introducing this new adaptive technique, there are always some parameters being updated. These parameters are used to extract and choose the most distinguished features. This keeps the models learning from the samples. Besides, for the objective function with the traditional $L_2$-norm, there exists some components from the regularization term, which can disturb the sampling gradients. Our goal is to make models continue learning the features equally. So we do not need the traditional $L_2$-norm. Instead, we replace the $L_2$-norm using weight-decay \cite{DBLP:journals/corr/abs-1711-05101}. This insight is called the ``AW" method. Table 3 shows the highest accuracy, which is achieved by each algorithm.

% \subsection{\textbf{Discussion}}

\begin{table}[]
\setlength{\tabcolsep}{16pt}
\begin{tabular}{c|c|c}
\hline
Methods         & Accuracy (\%)                 & \# Epoch \\
\hline
SIGN-SGD   & 63.96\%               & 166                        \\
SIGNUM      & 67.12\%               & 291                        \\
ADAM          & 68.62\%                & 288                         \\ %\hline
signADAM    & 67.97\%               & 227                        \\ %\hline
signADAM++ & 65.38\%             & 285                        \\ %\hline
ADAMW        & 69.11\%               & 291                           \\ %\hline
signADAM++-AW   & \textbf{72.31}\%               & 278  \\
\hline
\end{tabular}
\vspace{1mm}
\caption{Comparison of the accuracies and needed epochs of all the algorithms on CIFAR-100. It is clear that the accuracy of signADAM++ is significantly better than those of the other algorithms on CIFAR-100.}
\end{table}

\section{CONCLUSIONS}
In this paper, we first proposed an efficient sign-based gradient descent optimization method for non-convex objective functions. Our method combines the low computation complexity advantage of SIGN-SGD and the adaptive property of ADAM. In particular, our algorithms incorporated the sign operation into ADAM. Besides, we defined the confidence function to make the model learn features equally. The motivation is supported by experimental results. signADAM++ will not be constrained by the depth of neural networks and has some obvious advantages over some state-of-the-art algorithms including SIGN-SGD, SIGNUM and ADAM. Some intuitive analysis shows that signADAM++ has much faster convergence than ADAM, which is a popular optimization algorithm for deep learning. We analyzed the convergence properties of our algorithms and the experimental results also confirmed our analysis. Furthermore, signADAM++ suits our new optimization framework based on the confidence function. In conclusion, we found that signADAM++ is a robust algorithm for training various deep neural networks. We also proposed a framework on the confidence function and reveal the relationship between the loss landscape and feature learning.

\section*{Acknowledgments}
This work was supported by the Project supported the Foundation for Innovative Research Groups of the National Natural Science Foundation of China (No.\ 61621005), the Major Research Plan of the National Natural Science Foundation of China (Nos.\ 91438201 and 91438103), the National Natural Science Foundation of China (Nos.\ 61876220, 61876221, 61836009, U1701267, 61871310, 61573267, 61502369 and 61473215), the Program for Cheung Kong Scholars and Innovative Research Team in University (No.\ IRT\_15R53), the Fund for Foreign Scholars in University Research and Teaching Programs (the 111 Project) (No.\ B07048), and the Science Foundation of Xidian University (Nos.\ 10251180018 and 10251180019).

\bibliographystyle{ACM-Reference-Format}
\bibliography{sample-base}


\begin{thebibliography}{42}

%%% the Web address.

\ifx \showCODEN    \undefined \def \showCODEN     #1{\unskip}     \fi
\ifx \showDOI      \undefined \def \showDOI       #1{#1}\fi
\ifx \showISBNx    \undefined \def \showISBNx     #1{\unskip}     \fi
\ifx \showISBNxiii \undefined \def \showISBNxiii  #1{\unskip}     \fi
\ifx \showISSN     \undefined \def \showISSN      #1{\unskip}     \fi
\ifx \showLCCN     \undefined \def \showLCCN      #1{\unskip}     \fi
\ifx \shownote     \undefined \def \shownote      #1{#1}          \fi
\ifx \showarticletitle \undefined \def \showarticletitle #1{#1}   \fi
\ifx \showURL      \undefined \def \showURL       {\relax}        \fi
% invisible to TeX
\providecommand\bibfield[2]{#2}
\providecommand\bibinfo[2]{#2}
\providecommand\natexlab[1]{#1}
\providecommand\showeprint[2][]{arXiv:#2}

\bibitem[\protect\citeauthoryear{Alistarh, Grubic, Li, Tomioka, and
  Vojnovic}{Alistarh et~al\mbox{.}}{2017}]%
        {alistarh2017qsgd}
\bibfield{author}{\bibinfo{person}{Dan Alistarh}, \bibinfo{person}{Demjan
  Grubic}, \bibinfo{person}{Jerry Li}, \bibinfo{person}{Ryota Tomioka}, {and}
  \bibinfo{person}{Milan Vojnovic}.} \bibinfo{year}{2017}\natexlab{}.
\newblock \showarticletitle{QSGD: Communication-efficient SGD via gradient
  quantization and encoding}. In \bibinfo{booktitle}{\emph{NeurIPS}}.
  \bibinfo{pages}{1709--1720}.
\newblock


\bibitem[\protect\citeauthoryear{Attwell and Laughlin}{Attwell and
  Laughlin}{2001}]%
        {attwell2001energy}
\bibfield{author}{\bibinfo{person}{David Attwell} {and}
  \bibinfo{person}{Simon~B Laughlin}.} \bibinfo{year}{2001}\natexlab{}.
\newblock \showarticletitle{An energy budget for signaling in the grey matter
  of the brain}.
\newblock \bibinfo{journal}{\emph{Journal of Cerebral Blood Flow \&
  Metabolism}} \bibinfo{volume}{21}, \bibinfo{number}{10}
  (\bibinfo{year}{2001}), \bibinfo{pages}{1133--1145}.
\newblock


\bibitem[\protect\citeauthoryear{Balles and Hennig}{Balles and Hennig}{2017}]%
        {balles2017dissecting}
\bibfield{author}{\bibinfo{person}{Lukas Balles} {and} \bibinfo{person}{Philipp
  Hennig}.} \bibinfo{year}{2017}\natexlab{}.
\newblock \showarticletitle{Dissecting adam: The sign, magnitude and variance
  of stochastic gradients}.
\newblock \bibinfo{journal}{\emph{arXiv preprint arXiv:1705.07774}}
  (\bibinfo{year}{2017}).
\newblock


\bibitem[\protect\citeauthoryear{Bernstein, Azizzadenesheli, Wang, and
  Anandkumar}{Bernstein et~al\mbox{.}}{2018a}]%
        {bernstein2018convergence}
\bibfield{author}{\bibinfo{person}{Jeremy Bernstein}, \bibinfo{person}{Kamyar
  Azizzadenesheli}, \bibinfo{person}{Yu-Xiang Wang}, {and}
  \bibinfo{person}{Anima Anandkumar}.} \bibinfo{year}{2018}\natexlab{a}.
\newblock \showarticletitle{Convergence rate of sign stochastic gradient
  descent for non-convex functions}.
\newblock  (\bibinfo{year}{2018}).
\newblock


\bibitem[\protect\citeauthoryear{Bernstein, Wang, Azizzadenesheli, and
  Anandkumar}{Bernstein et~al\mbox{.}}{2018b}]%
        {bernstein2018signsgd}
\bibfield{author}{\bibinfo{person}{Jeremy Bernstein}, \bibinfo{person}{Yu-Xiang
  Wang}, \bibinfo{person}{Kamyar Azizzadenesheli}, {and} \bibinfo{person}{Anima
  Anandkumar}.} \bibinfo{year}{2018}\natexlab{b}.
\newblock \showarticletitle{signSGD: Compressed optimisation for non-convex
  problems}.
\newblock \bibinfo{journal}{\emph{arXiv preprint arXiv:1802.04434}}
  (\bibinfo{year}{2018}).
\newblock


\bibitem[\protect\citeauthoryear{Boyd and Vandenberghe}{Boyd and
  Vandenberghe}{2004}]%
        {boyd2004convex}
\bibfield{author}{\bibinfo{person}{Stephen Boyd} {and} \bibinfo{person}{Lieven
  Vandenberghe}.} \bibinfo{year}{2004}\natexlab{}.
\newblock \bibinfo{booktitle}{\emph{Convex optimization}}.
\newblock \bibinfo{publisher}{Cambridge university press}.
\newblock


\bibitem[\protect\citeauthoryear{De~Sa, Zhang, Olukotun, and R{\'e}}{De~Sa
  et~al\mbox{.}}{2015}]%
        {de2015taming}
\bibfield{author}{\bibinfo{person}{Christopher~M De~Sa}, \bibinfo{person}{Ce
  Zhang}, \bibinfo{person}{Kunle Olukotun}, {and} \bibinfo{person}{Christopher
  R{\'e}}.} \bibinfo{year}{2015}\natexlab{}.
\newblock \showarticletitle{Taming the wild: A unified analysis of
  hogwild-style algorithms}. In \bibinfo{booktitle}{\emph{NeurIPS}}.
  \bibinfo{pages}{2674--2682}.
\newblock


\bibitem[\protect\citeauthoryear{Donoho and Johnstone}{Donoho and
  Johnstone}{1994}]%
        {donoho1994ideal}
\bibfield{author}{\bibinfo{person}{David~L Donoho} {and}
  \bibinfo{person}{Jain~M Johnstone}.} \bibinfo{year}{1994}\natexlab{}.
\newblock \showarticletitle{Ideal spatial adaptation by wavelet shrinkage}.
\newblock \bibinfo{journal}{\emph{biometrika}} \bibinfo{volume}{81},
  \bibinfo{number}{3} (\bibinfo{year}{1994}), \bibinfo{pages}{425--455}.
\newblock


\bibitem[\protect\citeauthoryear{Douglas and Martin}{Douglas and
  Martin}{2007}]%
        {douglas2007recurrent}
\bibfield{author}{\bibinfo{person}{Rodney~J Douglas} {and}
  \bibinfo{person}{Kevan~AC Martin}.} \bibinfo{year}{2007}\natexlab{}.
\newblock \showarticletitle{Recurrent neuronal circuits in the neocortex}.
\newblock \bibinfo{journal}{\emph{Current biology}} \bibinfo{volume}{17},
  \bibinfo{number}{13} (\bibinfo{year}{2007}), \bibinfo{pages}{R496--R500}.
\newblock


\bibitem[\protect\citeauthoryear{Dozat}{Dozat}{2016}]%
        {dozat2016Nadam}
\bibfield{author}{\bibinfo{person}{Timothy Dozat}.}
  \bibinfo{year}{2016}\natexlab{}.
\newblock \showarticletitle{Incorporating Nesterov Momentum into Adam}. In
  \bibinfo{booktitle}{\emph{ICLR Workshop}}. \bibinfo{pages}{2013--2016}.
\newblock


\bibitem[\protect\citeauthoryear{Duchi, Hazan, and Singer}{Duchi
  et~al\mbox{.}}{2011}]%
        {duchi2011adaptive}
\bibfield{author}{\bibinfo{person}{John Duchi}, \bibinfo{person}{Elad Hazan},
  {and} \bibinfo{person}{Yoram Singer}.} \bibinfo{year}{2011}\natexlab{}.
\newblock \showarticletitle{Adaptive subgradient methods for online learning
  and stochastic optimization}.
\newblock \bibinfo{journal}{\emph{Journal of Machine Learning Research}}
  \bibinfo{volume}{12}, \bibinfo{number}{Jul} (\bibinfo{year}{2011}),
  \bibinfo{pages}{2121--2159}.
\newblock


\bibitem[\protect\citeauthoryear{Glorot, Bordes, and Bengio}{Glorot
  et~al\mbox{.}}{2011}]%
        {glorot2011deep}
\bibfield{author}{\bibinfo{person}{Xavier Glorot}, \bibinfo{person}{Antoine
  Bordes}, {and} \bibinfo{person}{Yoshua Bengio}.}
  \bibinfo{year}{2011}\natexlab{}.
\newblock \showarticletitle{Deep sparse rectifier neural networks}. In
  \bibinfo{booktitle}{\emph{AIStats}}. \bibinfo{pages}{315--323}.
\newblock


\bibitem[\protect\citeauthoryear{He, Zhang, Ren, and Sun}{He
  et~al\mbox{.}}{2016}]%
        {he2016deep}
\bibfield{author}{\bibinfo{person}{Kaiming He}, \bibinfo{person}{Xiangyu
  Zhang}, \bibinfo{person}{Shaoqing Ren}, {and} \bibinfo{person}{Jian Sun}.}
  \bibinfo{year}{2016}\natexlab{}.
\newblock \showarticletitle{Deep residual learning for image recognition}. In
  \bibinfo{booktitle}{\emph{Proceedings of the IEEE conference on computer
  vision and pattern recognition}}. \bibinfo{pages}{770--778}.
\newblock


\bibitem[\protect\citeauthoryear{Hinton, Osindero, and Teh}{Hinton
  et~al\mbox{.}}{2006}]%
        {hinton2006fast}
\bibfield{author}{\bibinfo{person}{Geoffrey~E Hinton}, \bibinfo{person}{Simon
  Osindero}, {and} \bibinfo{person}{Yee-Whye Teh}.}
  \bibinfo{year}{2006}\natexlab{}.
\newblock \showarticletitle{A fast learning algorithm for deep belief nets}.
\newblock \bibinfo{journal}{\emph{Neural computation}} \bibinfo{volume}{18},
  \bibinfo{number}{7} (\bibinfo{year}{2006}), \bibinfo{pages}{1527--1554}.
\newblock


\bibitem[\protect\citeauthoryear{Hu, Shen, and Sun}{Hu et~al\mbox{.}}{2018}]%
        {hu2018squeeze}
\bibfield{author}{\bibinfo{person}{Jie Hu}, \bibinfo{person}{Li Shen}, {and}
  \bibinfo{person}{Gang Sun}.} \bibinfo{year}{2018}\natexlab{}.
\newblock \showarticletitle{Squeeze-and-excitation networks}. In
  \bibinfo{booktitle}{\emph{Proceedings of the IEEE conference on computer
  vision and pattern recognition}}. \bibinfo{pages}{7132--7141}.
\newblock


\bibitem[\protect\citeauthoryear{Karimi, Nutini, and Schmidt}{Karimi
  et~al\mbox{.}}{2016}]%
        {karimi2016linear}
\bibfield{author}{\bibinfo{person}{Hamed Karimi}, \bibinfo{person}{Julie
  Nutini}, {and} \bibinfo{person}{Mark Schmidt}.}
  \bibinfo{year}{2016}\natexlab{}.
\newblock \showarticletitle{Linear convergence of gradient and
  proximal-gradient methods under the polyak-{\l}ojasiewicz condition}. In
  \bibinfo{booktitle}{\emph{ECML-PKDD}}. Springer, \bibinfo{pages}{795--811}.
\newblock


\bibitem[\protect\citeauthoryear{Kingma and Ba}{Kingma and Ba}{2014}]%
        {kingma2014adam}
\bibfield{author}{\bibinfo{person}{Diederik~P Kingma} {and}
  \bibinfo{person}{Jimmy Ba}.} \bibinfo{year}{2014}\natexlab{}.
\newblock \showarticletitle{Adam: A method for stochastic optimization}.
\newblock \bibinfo{journal}{\emph{arXiv preprint arXiv:1412.6980}}
  (\bibinfo{year}{2014}).
\newblock


\bibitem[\protect\citeauthoryear{Langford, Li, and Zhang}{Langford
  et~al\mbox{.}}{2009}]%
        {langford2009sparse}
\bibfield{author}{\bibinfo{person}{John Langford}, \bibinfo{person}{Lihong Li},
  {and} \bibinfo{person}{Tong Zhang}.} \bibinfo{year}{2009}\natexlab{}.
\newblock \showarticletitle{Sparse online learning via truncated gradient}.
\newblock \bibinfo{journal}{\emph{JMLR}} \bibinfo{volume}{10},
  \bibinfo{number}{Mar} (\bibinfo{year}{2009}), \bibinfo{pages}{777--801}.
\newblock


\bibitem[\protect\citeauthoryear{LeCun, Bengio, and Hinton}{LeCun
  et~al\mbox{.}}{2015}]%
        {lecun2015deep}
\bibfield{author}{\bibinfo{person}{Yann LeCun}, \bibinfo{person}{Yoshua
  Bengio}, {and} \bibinfo{person}{Geoffrey Hinton}.}
  \bibinfo{year}{2015}\natexlab{}.
\newblock \showarticletitle{Deep learning}.
\newblock \bibinfo{journal}{\emph{nature}} \bibinfo{volume}{521},
  \bibinfo{number}{7553} (\bibinfo{year}{2015}), \bibinfo{pages}{436}.
\newblock


\bibitem[\protect\citeauthoryear{Lennie}{Lennie}{2003}]%
        {lennie2003cost}
\bibfield{author}{\bibinfo{person}{Peter Lennie}.}
  \bibinfo{year}{2003}\natexlab{}.
\newblock \showarticletitle{The cost of cortical computation}.
\newblock \bibinfo{journal}{\emph{Current biology}} \bibinfo{volume}{13},
  \bibinfo{number}{6} (\bibinfo{year}{2003}), \bibinfo{pages}{493--497}.
\newblock


\bibitem[\protect\citeauthoryear{Livnat, Papadimitriou, Dushoff, and
  Feldman}{Livnat et~al\mbox{.}}{2008}]%
        {livnat2008mixability}
\bibfield{author}{\bibinfo{person}{Adi Livnat}, \bibinfo{person}{Christos
  Papadimitriou}, \bibinfo{person}{Jonathan Dushoff}, {and}
  \bibinfo{person}{Marcus~W Feldman}.} \bibinfo{year}{2008}\natexlab{}.
\newblock \showarticletitle{A mixability theory for the role of sex in
  evolution}.
\newblock \bibinfo{journal}{\emph{Proceedings of the National Academy of
  Sciences}} \bibinfo{volume}{105}, \bibinfo{number}{50}
  (\bibinfo{year}{2008}), \bibinfo{pages}{19803--19808}.
\newblock


\bibitem[\protect\citeauthoryear{Loshchilov and Hutter}{Loshchilov and
  Hutter}{2017a}]%
        {loshchilov2017fixing}
\bibfield{author}{\bibinfo{person}{Ilya Loshchilov} {and}
  \bibinfo{person}{Frank Hutter}.} \bibinfo{year}{2017}\natexlab{a}.
\newblock \showarticletitle{Fixing weight decay regularization in adam}.
\newblock \bibinfo{journal}{\emph{arXiv preprint arXiv:1711.05101}}
  (\bibinfo{year}{2017}).
\newblock


\bibitem[\protect\citeauthoryear{Loshchilov and Hutter}{Loshchilov and
  Hutter}{2017b}]%
        {DBLP:journals/corr/abs-1711-05101}
\bibfield{author}{\bibinfo{person}{Ilya Loshchilov} {and}
  \bibinfo{person}{Frank Hutter}.} \bibinfo{year}{2017}\natexlab{b}.
\newblock \showarticletitle{Fixing Weight Decay Regularization in Adam}.
\newblock \bibinfo{journal}{\emph{CoRR}}  \bibinfo{volume}{abs/1711.05101}
  (\bibinfo{year}{2017}).
\newblock
\showeprint[arxiv]{1711.05101}
\urldef\tempurl%
\url{http://arxiv.org/abs/1711.05101}
\showURL{%
\tempurl}


\bibitem[\protect\citeauthoryear{Luo, Xiong, Liu, and Sun}{Luo
  et~al\mbox{.}}{2019}]%
        {luo2019adaptive}
\bibfield{author}{\bibinfo{person}{Liangchen Luo}, \bibinfo{person}{Yuanhao
  Xiong}, \bibinfo{person}{Yan Liu}, {and} \bibinfo{person}{Xu Sun}.}
  \bibinfo{year}{2019}\natexlab{}.
\newblock \showarticletitle{Adaptive gradient methods with dynamic bound of
  learning rate}.
\newblock \bibinfo{journal}{\emph{arXiv preprint arXiv:1902.09843}}
  (\bibinfo{year}{2019}).
\newblock


\bibitem[\protect\citeauthoryear{Nair and Hinton}{Nair and Hinton}{2010}]%
        {nair2010rectified}
\bibfield{author}{\bibinfo{person}{Vinod Nair} {and}
  \bibinfo{person}{Geoffrey~E Hinton}.} \bibinfo{year}{2010}\natexlab{}.
\newblock \showarticletitle{Rectified linear units improve restricted boltzmann
  machines}. In \bibinfo{booktitle}{\emph{ICML}}. \bibinfo{pages}{807--814}.
\newblock


\bibitem[\protect\citeauthoryear{Qian}{Qian}{1999}]%
        {qian1999momentum}
\bibfield{author}{\bibinfo{person}{Ning Qian}.}
  \bibinfo{year}{1999}\natexlab{}.
\newblock \showarticletitle{On the momentum term in gradient descent learning
  algorithms}.
\newblock \bibinfo{journal}{\emph{Neural networks}} \bibinfo{volume}{12},
  \bibinfo{number}{1} (\bibinfo{year}{1999}), \bibinfo{pages}{145--151}.
\newblock


\bibitem[\protect\citeauthoryear{Reddi, Kale, and Kumar}{Reddi
  et~al\mbox{.}}{2018}]%
        {reddi2018convergence}
\bibfield{author}{\bibinfo{person}{Sashank~J Reddi}, \bibinfo{person}{Satyen
  Kale}, {and} \bibinfo{person}{Sanjiv Kumar}.}
  \bibinfo{year}{2018}\natexlab{}.
\newblock \showarticletitle{On the convergence of adam and beyond}.
\newblock  (\bibinfo{year}{2018}).
\newblock


\bibitem[\protect\citeauthoryear{Richt{\'a}rik and
  Tak{\'a}{\v{c}}}{Richt{\'a}rik and Tak{\'a}{\v{c}}}{2014}]%
        {richtarik2014iteration}
\bibfield{author}{\bibinfo{person}{Peter Richt{\'a}rik} {and}
  \bibinfo{person}{Martin Tak{\'a}{\v{c}}}.} \bibinfo{year}{2014}\natexlab{}.
\newblock \showarticletitle{Iteration complexity of randomized block-coordinate
  descent methods for minimizing a composite function}.
\newblock \bibinfo{journal}{\emph{Mathematical Programming}}
  \bibinfo{volume}{144}, \bibinfo{number}{1-2} (\bibinfo{year}{2014}),
  \bibinfo{pages}{1--38}.
\newblock


\bibitem[\protect\citeauthoryear{Riedmiller and Braun}{Riedmiller and
  Braun}{1993}]%
        {riedmiller1993direct}
\bibfield{author}{\bibinfo{person}{Martin Riedmiller} {and}
  \bibinfo{person}{Heinrich Braun}.} \bibinfo{year}{1993}\natexlab{}.
\newblock \showarticletitle{A direct adaptive method for faster backpropagation
  learning: The RPROP algorithm}. In \bibinfo{booktitle}{\emph{Proceedings of
  the IEEE international conference on neural networks}},
  Vol.~\bibinfo{volume}{1993}. San Francisco, \bibinfo{pages}{586--591}.
\newblock


\bibitem[\protect\citeauthoryear{Robbins and Monro}{Robbins and Monro}{1951}]%
        {robbins1951stochastic}
\bibfield{author}{\bibinfo{person}{Herbert Robbins} {and}
  \bibinfo{person}{Sutton Monro}.} \bibinfo{year}{1951}\natexlab{}.
\newblock \showarticletitle{A stochastic approximation method}.
\newblock \bibinfo{journal}{\emph{The annals of mathematical statistics}}
  (\bibinfo{year}{1951}), \bibinfo{pages}{400--407}.
\newblock


\bibitem[\protect\citeauthoryear{Ruder}{Ruder}{2016}]%
        {ruder2016overview}
\bibfield{author}{\bibinfo{person}{Sebastian Ruder}.}
  \bibinfo{year}{2016}\natexlab{}.
\newblock \showarticletitle{An overview of gradient descent optimization
  algorithms}.
\newblock \bibinfo{journal}{\emph{arXiv preprint arXiv:1609.04747}}
  (\bibinfo{year}{2016}).
\newblock


\bibitem[\protect\citeauthoryear{Rumelhart, Hinton, and Williams}{Rumelhart
  et~al\mbox{.}}{1985}]%
        {rumelhart1985learning}
\bibfield{author}{\bibinfo{person}{David~E Rumelhart},
  \bibinfo{person}{Geoffrey~E Hinton}, {and} \bibinfo{person}{Ronald~J
  Williams}.} \bibinfo{year}{1985}\natexlab{}.
\newblock \bibinfo{booktitle}{\emph{Learning internal representations by error
  propagation}}.
\newblock \bibinfo{type}{{T}echnical {R}eport}.
  \bibinfo{institution}{California Univ San Diego La Jolla Inst for Cognitive
  Science}.
\newblock


\bibitem[\protect\citeauthoryear{Rumelhart, Hinton, Williams,
  et~al\mbox{.}}{Rumelhart et~al\mbox{.}}{1988}]%
        {rumelhart1988learning}
\bibfield{author}{\bibinfo{person}{David~E Rumelhart},
  \bibinfo{person}{Geoffrey~E Hinton}, \bibinfo{person}{Ronald~J Williams},
  {et~al\mbox{.}}} \bibinfo{year}{1988}\natexlab{}.
\newblock \showarticletitle{Learning representations by back-propagating
  errors}.
\newblock \bibinfo{journal}{\emph{Cognitive modeling}} \bibinfo{volume}{5},
  \bibinfo{number}{3} (\bibinfo{year}{1988}), \bibinfo{pages}{1}.
\newblock


\bibitem[\protect\citeauthoryear{Seide, Fu, Droppo, Li, and Yu}{Seide
  et~al\mbox{.}}{2014}]%
        {seide20141}
\bibfield{author}{\bibinfo{person}{Frank Seide}, \bibinfo{person}{Hao Fu},
  \bibinfo{person}{Jasha Droppo}, \bibinfo{person}{Gang Li}, {and}
  \bibinfo{person}{Dong Yu}.} \bibinfo{year}{2014}\natexlab{}.
\newblock \showarticletitle{1-bit stochastic gradient descent and its
  application to data-parallel distributed training of speech dnns}. In
  \bibinfo{booktitle}{\emph{Interspeech}}.
\newblock


\bibitem[\protect\citeauthoryear{Simonyan and Zisserman}{Simonyan and
  Zisserman}{2014}]%
        {simonyan2014very}
\bibfield{author}{\bibinfo{person}{Karen Simonyan} {and}
  \bibinfo{person}{Andrew Zisserman}.} \bibinfo{year}{2014}\natexlab{}.
\newblock \showarticletitle{Very deep convolutional networks for large-scale
  image recognition}.
\newblock \bibinfo{journal}{\emph{arXiv preprint arXiv:1409.1556}}
  (\bibinfo{year}{2014}).
\newblock


\bibitem[\protect\citeauthoryear{Sutton}{Sutton}{1986}]%
        {sutton1986two}
\bibfield{author}{\bibinfo{person}{Richard Sutton}.}
  \bibinfo{year}{1986}\natexlab{}.
\newblock \showarticletitle{Two problems with back propagation and other
  steepest descent learning procedures for networks}. In
  \bibinfo{booktitle}{\emph{Proceedings of the Eighth Annual Conference of the
  Cognitive Science Society, 1986}}. \bibinfo{pages}{823--832}.
\newblock


\bibitem[\protect\citeauthoryear{Szegedy, Liu, Jia, Sermanet, Reed, Anguelov,
  Erhan, Vanhoucke, and Rabinovich}{Szegedy et~al\mbox{.}}{2015}]%
        {szegedy2015going}
\bibfield{author}{\bibinfo{person}{Christian Szegedy}, \bibinfo{person}{Wei
  Liu}, \bibinfo{person}{Yangqing Jia}, \bibinfo{person}{Pierre Sermanet},
  \bibinfo{person}{Scott Reed}, \bibinfo{person}{Dragomir Anguelov},
  \bibinfo{person}{Dumitru Erhan}, \bibinfo{person}{Vincent Vanhoucke}, {and}
  \bibinfo{person}{Andrew Rabinovich}.} \bibinfo{year}{2015}\natexlab{}.
\newblock \showarticletitle{Going deeper with convolutions}. In
  \bibinfo{booktitle}{\emph{Proceedings of the IEEE conference on computer
  vision and pattern recognition}}. \bibinfo{pages}{1--9}.
\newblock


\bibitem[\protect\citeauthoryear{Tieleman and Hinton}{Tieleman and
  Hinton}{2012}]%
        {Tieleman2012}
\bibfield{author}{\bibinfo{person}{T. Tieleman} {and} \bibinfo{person}{G.
  Hinton}.} \bibinfo{year}{2012}\natexlab{}.
\newblock \bibinfo{title}{{Lecture 6.5---RmsProp: Divide the gradient by a
  running average of its recent magnitude}}.
\newblock \bibinfo{howpublished}{COURSERA: Neural Networks for Machine
  Learning}.
\newblock


\bibitem[\protect\citeauthoryear{Tishby and Zaslavsky}{Tishby and
  Zaslavsky}{2015}]%
        {tishby2015deep}
\bibfield{author}{\bibinfo{person}{Naftali Tishby} {and} \bibinfo{person}{Noga
  Zaslavsky}.} \bibinfo{year}{2015}\natexlab{}.
\newblock \showarticletitle{Deep learning and the information bottleneck
  principle}. In \bibinfo{booktitle}{\emph{2015 IEEE Information Theory
  Workshop (ITW)}}. IEEE, \bibinfo{pages}{1--5}.
\newblock


\bibitem[\protect\citeauthoryear{Wen, Xu, Yan, Wu, Wang, Chen, and Li}{Wen
  et~al\mbox{.}}{2017}]%
        {wen2017terngrad}
\bibfield{author}{\bibinfo{person}{Wei Wen}, \bibinfo{person}{Cong Xu},
  \bibinfo{person}{Feng Yan}, \bibinfo{person}{Chunpeng Wu},
  \bibinfo{person}{Yandan Wang}, \bibinfo{person}{Yiran Chen}, {and}
  \bibinfo{person}{Hai Li}.} \bibinfo{year}{2017}\natexlab{}.
\newblock \showarticletitle{Terngrad: Ternary gradients to reduce communication
  in distributed deep learning}. In \bibinfo{booktitle}{\emph{NeurIPS}}.
  \bibinfo{pages}{1509--1519}.
\newblock


\bibitem[\protect\citeauthoryear{Zeiler}{Zeiler}{2012}]%
        {zeiler2012adadelta}
\bibfield{author}{\bibinfo{person}{Matthew~D Zeiler}.}
  \bibinfo{year}{2012}\natexlab{}.
\newblock \showarticletitle{ADADELTA: an adaptive learning rate method}.
\newblock \bibinfo{journal}{\emph{arXiv preprint arXiv:1212.5701}}
  (\bibinfo{year}{2012}).
\newblock


\bibitem[\protect\citeauthoryear{Zhang, Zhang, Li, and Qiao}{Zhang
  et~al\mbox{.}}{2016}]%
        {7553523}
\bibfield{author}{\bibinfo{person}{K. Zhang}, \bibinfo{person}{Z. Zhang},
  \bibinfo{person}{Z. Li}, {and} \bibinfo{person}{Y. Qiao}.}
  \bibinfo{year}{2016}\natexlab{}.
\newblock \showarticletitle{Joint Face Detection and Alignment Using Multitask
  Cascaded Convolutional Networks}.
\newblock \bibinfo{journal}{\emph{IEEE Signal Processing Letters}}
  \bibinfo{volume}{23}, \bibinfo{number}{10} (\bibinfo{date}{Oct}
  \bibinfo{year}{2016}), \bibinfo{pages}{1499--1503}.
\newblock
\showISSN{1070-9908}
\urldef\tempurl%
\url{https://doi.org/10.1109/LSP.2016.2603342}
\showDOI{\tempurl}


\end{thebibliography}

\section*{APPENDIX A: Proof of Lemma 1}
\begin{proof}
$$\forall i,\;\;m_{1,i}=0,  \:\:\:m_{2,i}\in\{\pm (1-\beta) , 0\}.$$
Since $\beta \in (0,1)$, then $|m_{2,i}|\leqslant 1$.
Assume $\forall i,|m_{k,i}|\leqslant 1$, then we have
\begin{align*}
\forall i,|m_{k+1,i}|&=|\beta\cdot m_{k,i} +(1-\beta)sign(\tilde{g}_{k,i})|\\
&\leqslant \beta\cdot |m_{k,i}| +(1-\beta)\cdot|sign(\tilde{g}_{k,i})|\\
&\leqslant \beta +1-\beta =1.
\end{align*}
This completes the proof.
\end{proof}

\section*{APPENDIX B: Proof of Lemma 2}
\begin{proof}
Using Lemma F.3 in SIGNSGD~\cite{bernstein2018signsgd}, then we can have the following conclusion:
Under Assumption 2, for any $s \in R^d$, $(\left | s_i\leq 1 \right |$, $i=1,2,\cdots,d )$, any x $\in R^d $ and any $ \epsilon \leq \delta$, then
\begin{align*}
\|g(x+\epsilon s)-g(x)\|_{1} &\leq 2 \epsilon\|\vec{L}\|_{1}.
\end{align*}
We also have
\begin{equation*}
\begin{split}
\|g_{k-t}-g_{k}\|_{1} & \leq \sum_{l=0}^{t-1}\left\|g_{k-l}-g_{k-l-1}\right\|_{1} \\
& \leq 2\|\vec{L}\|_{1} \sum_{l=0}^{t-1} \delta_{k-l-1} \leq \sum_{l=0}^{t-1} \frac{2\|\vec{L}\|_{1} \delta}{\sqrt{k-l}} \\ & \leq 2\|\vec{L}\|_{1} \delta \int_{k-t}^{k} \frac{d x}{\sqrt{x}}=4\|\vec{L}\|_{1} \delta(\sqrt{k}-\sqrt{k-t}) \\ & \leq 4\|\vec{L}\|_{1} \delta \sqrt{k}\left(1-\sqrt{1-\frac{t}{k}}\right) \\ & \leq 4\|\vec{L}\|_{1} \delta \frac{t}{\sqrt{k}}.
\end{split}
\end{equation*}
Let $\delta=\frac{1}{\sqrt{\|\vec{L}\|_{1}}}$, then
\begin{equation*}
\begin{split}
\frac{1-\beta}{1-\beta^k}\sum^{K-1}_{t=0}\beta^t[||g_{k-t} - g_{k}||_1]&\leq
\frac{1-\beta}{1-\beta}\sum^{\infty}_{t=0}\beta^t[||g_{k-t} - g_{k}||_1] \\
&\leq 4\frac{\sqrt{\left \| \vec{L} \right \|_{1}}}{\sqrt{k}}\frac{\beta}{(1-\beta)^{2}}.
\end{split}
\end{equation*}
This completes the proof.
\end{proof}

\section*{APPENDIX C: Proof of Theorem 1}
\begin{proof}
Suppose Assumptions 1, 2 and 3 hold. Following the proof non-convex SGD, we modify Assumption 2 as follows.
\begin{align*}
f_{k+1}-f_{k}&\leqslant {g_k}^T  (x_{k+1}-x_k)+\sum_{i=1}^{d} \frac{L_i}{2} (x_{k+1}-x_k) _{i}^2\\
&=-\delta_k  {g_k}^T    m_k + {\delta_k}^2 \sum _{i=1}^{d}\frac{L_i}{2}  {m_{k,i}}^2.
\end{align*}
Then we apply Lemma1 (i.e., $|m_{k,i}|\leqslant1$) into the above inequality and can get
\begin{align*}
f_{k+1}-f_{k}&\leqslant -\delta_k  {g_k}^T    m_k + {\delta_k}^2 \sum _{i=1}^{d}\frac{L_i}{2}.
\end{align*}
Now we calculate the mathematical expectation of both sides of the above inequality.

\onecolumn
\begin{align*}
\mathbb{E}[f_{k+1}\!-\!f_{k}|x_k]&\leqslant -\delta_k \mathbb{E}[{g_k}^T  m_k]+\frac{\delta_k^2}{2}  \left \| \vec{L} \right \|_1\\
&= -\delta_k\mathbb{E}[{g_k}^T \frac{1-\beta}{1-\beta^k}\sum^{k-1}_{t=0}{\beta^t sign(\tilde{g}_{k-t})} ]+\frac{\delta_k^2}{2}  \left \| \vec{L} \right \|_1\\
&= -\delta_k \frac{1-\beta}{1-\beta^k}\sum^{k-1}_{t=0}\mathbb{E}\left\{\beta^t\left[\|g_k\|_1 - 2\sum^{d}_{i=1}|g_{k,i}|\mathbb{I}(sign(g_{k,i}\neq sign(\tilde{g}_{k-t,i}))\right]\right\}
 + \frac{\delta_k^2}{2}  \left \| \vec{L} \right \|_1\\
&= -\delta_k \frac{1-\beta}{1-\beta^k}\sum^{k-1}_{t=0}\left(\beta^t\left[\|g_k\|_1 -2\sum^{d}_{i=1}|g_{k,i}|\mathbb{P}(sign(g_{k,i}\neq sign(\tilde{g}_{k-t,i}))\right]\right)
  + \frac{\delta_k^2}{2}  \left \| \vec{L} \right \|_1.
\end{align*}
\begin{align*}
\mathbb{E}[f_{k+1}-f_{k}|x_k]&\leqslant
-\delta_k \left \| g_k \right \|_1 + 2\delta_k \frac{1-\beta}{1-\beta^k}\sum^{k-1}_{t=0}\beta^t\sum _{i=1}^{d}|g_{k,i}|\mathbb{P}[|\tilde{g}_{k-t,i}-g_{k,i}| \geqslant g_{k,i}] + \frac{\delta_k^2}{2}  \left \| \vec{L} \right \|_1\\
&\leqslant -\delta_k \| g_k \|_1 + 2\delta_k \frac{1-\beta}{1-\beta^k}\sum^{k-1}_{t=0}\beta^t\sum _{i=1}^{d}\mathbb{E}[|\tilde{g}_{k-t,i} - g_{k,i}|] +\frac{\delta_k^2}{2}  \left \| \vec{L}\right \|_1\\
&\leqslant -\delta_k \| g_k \|_1 + 2\delta_k \frac{1-\beta}{1-\beta^k}\sum^{k-1}_{t=0}\beta^t\sum _{i=1}^{d}\left(\mathbb{E}[|\tilde{g}_{k-t,i} - g_{k-t,i}|] + \mathbb{E}[|g_{k-t,i} - g_{k,i}|]\right)
+\frac{\delta_k^2}{2}  \left \| \vec{L} \right \|_1\\
&\leqslant -\delta_k \| g_k \|_1 + 2\delta_k \frac{1-\beta}{1-\beta^k}\sum^{k-1}_{t=0}\beta^t\sum _{i=1}^{d} \left(\sqrt{\mathbb{E}[|\tilde{g}_{k-t,i} - g_{k-t,i}|^2]} + \mathbb{E}[|g_{k-t,i} - g_{k,i}|]\right)
+\frac{\delta_k^2}{2}  \left \| \vec{L} \right \|_1\\
&\leqslant -\delta_k \| g_k \|_1 + 2\delta_k \frac{1-\beta}{1-\beta^k}\sum^{k-1}_{t=0}\beta^t\frac{\|\vec{\sigma}\|_1}{\sqrt{n_k}}
+2\delta_k \frac{1-\beta}{1-\beta^k}\sum^{k-1}_{t=0}\beta^t[||g_{k-t} - g_{k}||_1]
+\frac{\delta_k^2}{2}  \left \| \vec{L} \right \|_1.
\end{align*}
Note that $\mathbb{I}\left( \cdot\right)=1$ when $\left( \cdot\right)$ holds, otherwise $\mathbb{I}\left( \cdot\right)=0$. $\mathbb{P}\left( \cdot\right)$ stands for the probability of $\left( \cdot\right)$.

Using Lemma 2, we can obtain
\begin{align*}
\mathbb{E}[f_{k+1}\!-\!f_{k}|x_k]&\leqslant -\delta_k \| g_k \|_1 + 2\delta_k\left( \frac{\|\vec{\sigma}\|_1}{\sqrt{n_k}}  + \frac{1-\beta}{1-\beta^k}\sum^{k-1}_{t=0}\beta^t[||g_{k-t} - g_{k}||_1]\right)+\frac{\delta_k^2}{2}  \left \| \vec{L} \right \|_1\\
&\leqslant -\frac{1}{\sqrt{\|\vec{L}\|_1 K}} \| g_k \|_1 +
\frac{2\|\vec{\sigma}\|_1}{\sqrt{\|\vec{L}\|_1} K} + \frac{1}{K}\frac{8\beta}{(1-\beta)^2}
+\frac{1}{2K}.
\end{align*}
Since
\begin{align*}
f(\theta_{0})-f(\theta^{\ast}) & \geq f(\theta_{0})-\mathbb{E}\left[f_{K}\right]\\
&=\mathbb{E}\left[\sum_{k=0}^{K-1} f_{k}-f_{k+1}\right] \\
&\geq \mathbb{E} \sum_{k=0}^{K-1}\left[\frac{1}{\sqrt{\|\vec{L}\|_{1} K}}\|g_{k}\|_{1}-\frac{1}{2 \sqrt{\|\vec{L}\|_{1}} K}\left({ 4\|\vec\sigma\|_{1}+\sqrt{\|\vec{L}\|_{1}}}
+\frac{16\beta\sqrt{\|\vec{L}\|_{1}}}{(1-\beta)^2}\right)\right] \\
&=\sqrt{\frac{K}{\|\vec{L}\|_{1}}} \mathbb{E}\left[\frac{1}{K} \sum_{k=0}^{K-1}\left\|g_{k}\right\|_{1}\right]-\frac{1}{2 \sqrt{\|\vec{L}\|_{1}}}\left(4\|\vec\sigma\|_{1}+\sqrt{\|\vec{L}\|_{1}}
+\frac{16\beta\sqrt{\|\vec{L}\|_{1}}}{(1-\beta)^2}\right),
\end{align*}
then
\begin{align*}
\mathbb{E}\left[\frac{1}{K}\sum^{K-1}_{k=0}\|g_k\|_1\right]\leqslant \frac{1}{\sqrt{K}}\left[\sqrt{\|\vec{L}\|_{1}}\left(f(\theta_{0})-f(\theta^{\ast}) +\frac{1}{2}+\frac{8\beta}{(1-\beta)^2}\right)+2\|\vec\sigma\|_{1}\right].
\end{align*}
To square both sides of the above inequality, and we complete the proof.
\end{proof}

\end{document}